\documentclass[journal]{IEEEtran}
\usepackage[switch]{lineno}
\usepackage{amsmath,amsfonts}
\usepackage{algorithmic}
\usepackage{algorithm}
\usepackage{array}
\usepackage[caption=false,font=normalsize,labelfont=sf,textfont=sf]{subfig}
\usepackage{textcomp}
\usepackage{stfloats}
\usepackage{url}
\usepackage{verbatim}
\usepackage{graphicx}
\usepackage{cite}
\usepackage{multirow}
\usepackage{booktabs}
\usepackage{amsbsy}
\usepackage{rotating}
\usepackage{bigstrut}
\usepackage{color}
\usepackage[table]{xcolor}
\usepackage{tikz,orcidlink}
\hypersetup{
	colorlinks=true,
	linkcolor=black,
	citecolor=black
}

\let\oldequation\equation
\let\oldendequation\endequation
\renewenvironment{equation}{\linenomathNonumbers\oldequation}{\oldendequation\endlinenomath}

\makeatletter
\def\ps@IEEEtitlepagestyle{
	\def\@oddfoot{\mycopyrightnotice}
	\def\@evenfoot{}
}
\def\mycopyrightnotice{
	{\footnotesize
		\begin{minipage}{\textwidth}
			\centering
			Copyright~\copyright~2024 IEEE. Personal use of this material is permitted. However, permission to use this  \\ 
			material for any other purposes must be obtained from the IEEE by sending a request to pubs-permissions@ieee.org.
		\end{minipage}
	}
}
\begin{document}
	\title{Learning Adaptive Fusion Bank for Multi-modal Salient Object Detection}
	
	\author{Kunpeng Wang\hspace{-1.5mm}$^{~\orcidlink{0000-0002-2788-7583}}$, Zhengzheng Tu\hspace{-1.5mm}$^{~\orcidlink{0000-0002-9689-8657}}$, Chenglong Li\hspace{-1.5mm}$^{~\orcidlink{0000-0002-7233-2739}}$, Cheng Zhang, Bin Luo\hspace{-1.5mm}$^{~\orcidlink{0000-0001-5948-5055}}$,~\IEEEmembership{Senior~Member,~IEEE} 
		\thanks{This work was supported in part by the National Natural Science Foundation
			of China under Grant 62376005, 61876002, in part by the Natural Science Foundation of
			Anhui Higher Education Institution of China under Grant KJ2020A0033,
			in part by Anhui Provincial Natural Science foundation under Grant
			2108085MF211, in part by Anhui Energy Internet Joint Fund Project under
			Grant 2008085UD07, in part by Anhui Provincial Key Research
			and Development Program under Grant 202104d07020008. (Corresponding author is Bin Luo)}
		\thanks{Kunpeng Wang, Zhengzheng Tu, Cheng Zhang and Bin Luo are with 
			Information Materials
			and Intelligent Sensing Laboratory of Anhui Province, Anhui
			Provincial Key Laboratory of Multimodal Cognitive Computation,
			School of Computer Science and Technology, Anhui University, Hefei
			230601, China (e-mail: kp.wang@foxmail.com; zhengzhengahu@163.com; 
			cheng.zhang@ahu.edu.cn and luobin@ahu.edu.cn)}
		\thanks{Chenglong Li is with Anhui Provincial Key Laboratory of Multimodal
			Cognitive Computation, School of Artificial Intelligence, Anhui University,
			Hefei 230601, China, and also with the Institute of Physical Science and
			Information Technology, Anhui University, Hefei 230601, China (e-mail:
			lcl1314@foxmail.com)}}
	
	\markboth{IEEE TRANSACTIONS ON CIRCUITS and SYSTEMS for VIDEO TECHNOLOGY}%
	{Shell \MakeLowercase{\textit{et al.}}: A Sample Article Using IEEEtran.cls for IEEE Journals}
	
	\maketitle
	
	\begin{abstract}
		Multi-modal salient object detection (MSOD) aims to boost saliency detection performance by integrating visible sources with depth or thermal infrared ones. Existing methods generally design different fusion schemes to handle certain issues or challenges. Although these fusion schemes are effective at addressing specific issues or challenges, they may struggle to handle multiple complex challenges simultaneously. To solve this problem, we propose a novel adaptive fusion bank that makes full use of the complementary benefits from a set of basic fusion schemes to handle different challenges simultaneously for robust MSOD. We focus on handling five major challenges in MSOD, namely center bias, scale variation, image clutter, low illumination, and thermal crossover or depth ambiguity. The fusion bank proposed consists of five representative fusion schemes, which are specifically designed based on the characteristics of each challenge, respectively.
		The bank is scalable, and more fusion schemes could be incorporated into the bank for more challenges. To adaptively select the appropriate fusion scheme for multi-modal input, we introduce an adaptive ensemble module that forms the adaptive fusion bank, which is embedded into hierarchical layers for sufficient fusion of different source data. Moreover, we design an indirect interactive guidance module to accurately detect salient hollow objects via the skip integration of high-level semantic information and low-level spatial details. Extensive experiments on three RGBT datasets and seven RGBD datasets demonstrate that the proposed method achieves the outstanding performance compared to the state-of-the-art methods. The code and results are available at \href{https://github.com/Angknpng/LAFB}{https://github.com/Angknpng/LAFB}.
	\end{abstract}
	
	
	\section{Introduction}
	\begin{figure}[t]
		\centering
		\includegraphics[width=\columnwidth]{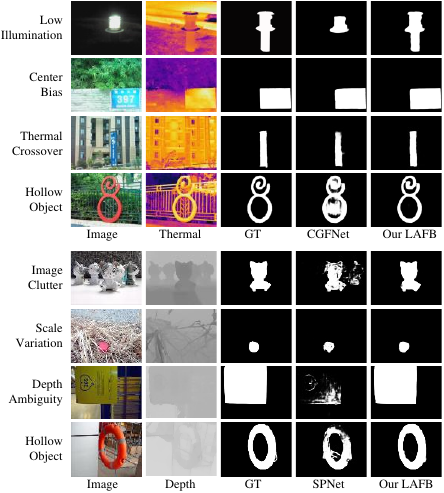}
		\caption{$\textbf{Top}$:  Comparison of our LAFB against a 
			state-of-the-art RGBT SOD method, e.g., 
			CGFNet~\cite{wang2021cgfnet}.
			$\textbf{Bottom}$: Comparison of our LAFB against a 
			state-of-the-art RGBD SOD method, e.g., 
			SPNet~\cite{zhou2021specificity}. The results show that 
			our LAFB can simultaneously handle both modality-shared challenges (i.e., center bias, scale variation, and image clutter) and modality-specific challenges (i.e., low illumination, thermal crossover or depth ambiguity), as well as accurately segment hollow objects. The compared methods can only tackle the certain challenge. For example, CGFNet can handle the center bias challenge well but fails on the low illumination and thermal crossover challenges, and SPNet can handle the scale variation challenge well but fails on the image clutter and depth ambiguity challenges. Neither of them can cope with the hollow objects.}
		
		\label{fig::branch}
	\end{figure}
	\IEEEPARstart{T}{he} purpose of salient object detection (SOD) is to detect and segment the most prominent region in a given visible image. Due to the practicality of SOD in image processing, it has become an auxiliary tool for many computer vision applications, such as video segmentation~\cite{wang2017saliency}, semantic segmentation~\cite{wang2022looking}, and 
	cropping~\cite{wang2018deep}. Most of SOD methods~\cite{zheng2021weakly,cong2022weakly,zhu2019aggregating,zhang2022progressive} are 
	based only on RGB images and achieve remarkable results. However, these RGB-based SOD methods may struggle 
	to segment salient objects from background in complex scenes, such as those with low illumination or similar foreground and background. To overcome the limitations of the RGB modality, depth maps and thermal infrared images are introduced to provide complementary information for multi-modal salient object detection (MSOD), including RGB-Depth (RGBD) SOD and RGB-Thermal (RGBT) SOD.
	
	Traditional MSOD methods~\cite{wang2018rgb,tang2019rgbt,tu2019rgb} rely on 
	manual features or heuristic priors to predict salient regions and outperform RGB-based SOD methods. However, they do not sufficiently explore the complementarity of multi-modal features or capture enough semantic information. Recently, many convolutional neural network-based methods~\cite{zhang2019rgb,tu2020rgbt,li2023mutual,tang2022hrtransnet,chen2022modality}
	are proposed to fully integrate multi-modal and multi-level features for accurate detection. Nevertheless, two key issues in MSOD remain to be addressed. (1) Existing multi-modal SOD methods~\cite{tu2020rgbt,zhang2020revisiting,huo2021efficient,qu2017rgbd,zhang2021uncertainty} mainly design different fusion strategies to exploit the complementary information between the modalities. Despite the performance improvement gained over RGB-based SOD methods, they still fail in some challenging scenarios, such as those with depth ambiguity, thermal crossover, low illumination, image clutter, etc. Some methods~\cite{wang2021cgfnet,liao2022cross,zhou2021ecffnet,liao2022cross,zhao2019contrast,piao2019depth,fan2020rethinking,li2021hierarchical} are recently proposed to address specific challenges, such as depth ambiguity~\cite{liao2022cross,zhao2019contrast,piao2019depth,fan2020rethinking,li2021hierarchical} and low illumination~\cite{wang2021cgfnet,liao2022cross,zhou2021ecffnet}. However, these fusion strategies aim to address certain challenges rather than addressing multiple challenges at the same time. 
	As shown in the upper three rows of both RGBT and RGBD cases in Fig.~\ref{fig::branch}, while the compared methods handle the center bias or scale variation challenges well, they fail in other main challenges. 
	(2) Multi-level fusion is widely used in multi-modal SOD methods~\cite{zhou2021specificity,wang2021cgfnet,Sun2021DeepRS,Ji_2021_DCF,liu2021visual,zhou2021ccafnet,mohammadi2020cagnet,zhang2020feature} to segment foreground from background distinctly by integrating high-level sematic features and low-level detailed features. The most common approach is top-down fusion, which directly combines the features of adjacent layers through concatenation, summation~\cite{wang2021cgfnet,Ji_2021_DCF,zhou2021ccafnet} or an integration module~\cite{zhou2021specificity,liu2021visual}. Due to the limited number of adjacent layers, these methods restrict the feature variety and prevent the full fusion of high-level and low-level features. To address this limitation, some other methods~\cite{Sun2021DeepRS,zhang2020feature} integrate multi-level features directly before top-down fusion to learn richer feature representations for each layer. However, these methods may overlook the unique characteristics of the features in different layers, limiting the exploration of complementarity between high-level and low-level features. Moreover, direct integration of multi-level features may compromise the original information in these features. Therefore, fully integrating multi-level features remains a key issue in MSOD. In complex scenes with hollow objects, such as those illustrated in the last row of both RGBT and RGBD cases in Fig.~\ref{fig::branch}, foreground and background are intertwined, making it difficult to accurately separate salient objects from the background. The compared methods~\cite{wang2021cgfnet,zhou2021specificity} struggle to distinguish the background inside the hollow object and misjudge non-salient regions as salient regions. This is mainly because they fail to fully aggregate semantic and detailed information.

	To address these issues, we propose to learn an adaptive fusion bank (LAFB) 
		for MSOD. To tackle the first issue, we design an adaptive fusion bank that consists of five representative fusion schemes and an adaptive ensemble module. The fusion schemes are separately designed according to the characteristics of the five major challenges in MSOD, namely center bias, scale variation, image clutter, low illumination, and thermal crossover or depth ambiguity. By embedding them into the encoder, the features extracted by the backbone can be decoupled into multiple specific features that deal with different challenges. The fusion schemes are simple but effective, and each of them only needs to learn a few parameters for a specific challenge and has less dependence on data. Additionally, the adaptive ensemble module takes the feature maps generated by different fusion schemes as input and learns their weights for adaptive fusion. In this way, our adaptive fusion bank can select the corresponding fusion schemes based on the challenges in different input data. Although the adaptive ensemble module is inspired by channel attention mechanism~\cite{woo2018cbam,hu2018squeeze}, their impact is completely different. Channel attention mainly focuses on refining feature representations by modeling the channel dependencies of independent features, while our adaptive ensemble module is responsible for adaptively integrating features that can deal with different challenges with different weights. In this way, the parameters of each fusion scheme can be trained accordingly. As shown in Fig.~\ref{fig::branch}, our method can better handle multiple complex challenges simultaneously due to the benefits from the adaptive fusion bank. 
	It is also worth noting that the bank is scalable and can incorporate more fusion schemes to deal with more challenges. In the future work, we will also introduce a common fusion scheme for the remaining challenges in MSOD. 
	
	For the second issue, we design an indirect interactive guidance 
	module (IIGM) which takes into account the unique characteristics of multi-level features to achieve full and smooth integration. 
	Considering direct integration may prevent the full integration of high-level and low-level features, we group the features of three adjacent layers and use the middle-level features as a medium to indirectly interact the high-level and low-level features.
	As the shown in the saliency maps in the last row of Fig.~\ref{fig::branch}, the 
	features through our IIGM can accurately segment the 
	salient regions of hollow objects. To sum up, our contributions 
	are as follows:
	\begin{itemize}
		\item We propose a novel adaptive fusion bank (AFB) to  
		handle multiple complex challenges in MSOD simultaneously. The AFB takes full
		advantage of the complementary benefits of different fusion schemes and 
		integrates them via an adaptive ensemble module. 
		The bank is scalable, and more fusion schemes can be 
		incorporated to address additional challenges.
		
		\item We design an indirect interactive guidance module (IIGM)
		to effectively integrate high-level semantic features and low-level 
		detailed features, allowing for precise segmentation of salient regions of hollow objects.
		
		\item Extensive experiments on seven RGBD SOD datasets and three RGBT SOD datasets as well as ablation studies demonstrate the effectiveness of our method and each component.
	\end{itemize}
	
	\section{Related Work}
	\subsection{RGBT Salient Object Detection}
	RGBT salient object detection (SOD) aims to detect common salient objects from RGB and thermal image pairs. Traditional 
	methods often use manual features~\cite{wang2018rgb} or 
	heuristic priors~\cite{tang2019rgbt,tu2019rgb} to discover complementary 
	information between modalities. While these have improved performance, they often fail to extract semantic information, making them less effective at dealing with complex challenges such as image cluster and low illumination.
	
	As convolutional neural network illustrates superior feature extraction ability, a large number of methods with different fusion modules have been proposed to address corresponding issues, such as modality complementarity~\cite{zhang2019rgb,tu2021multi,huo2021efficient} and misalignment~\cite{tu2022weakly}. Zhang et al.~\cite{zhang2019rgb} treat RGBT saliency detection as a feature fusion problem and propose a multi-modal fusion module to fully exploit the complementarity between RGB and thermal features. Tu et al.~\cite{tu2021multi} integrate local detail information, global context, and multi-modal features to model multiple interactions between two modalities. Huo et al.~\cite{huo2021efficient} propose a lightweight fusion module that efficiently integrates multi-modal features by exploiting the characteristics of two modalities. For unaligned RGBT image pairs, DCNet~\cite{tu2022weakly} establishes strong modality correlations through multi-branch affine transformations. However, due to the lack of the design specific for challenges, these methods might fail when facing challenging scenarios.
	
	Besides, some other methods are proposed to improve detection performance in challenging scenarios, such as low illumination~\cite{wang2021cgfnet,liao2022cross,zhou2021ecffnet} and thermal crossover~\cite{liao2022cross}. Wang et al.~\cite{wang2021cgfnet} fully integrate cross-modal information through mutual guidance between RGB and thermal modalities to improve performance in low illumination. ECFFNet~\cite{zhou2021ecffnet} designs a bilateral fusion of foreground and background information to extract complete boundaries, and then embeds them into multi-level features for accurate detection in low-light conditions. CCFENet~\cite{liao2022cross} embeds cross-modal fusion into the encoder to discriminatively extract features in the case of modality defects. 
	
	Although these studies are proposed for specific challenges, they might not cover multiple major challenges. To effectively deal with specific issues or challenges, the fusion schemes in the above methods are complex and involve a lot of parameters. However, due to the limited amount of multi-modal data, it is difficult to train these fusion schemes to learn a large number of parameters for multiple challenges. Instead, they may only be effective when learning a relatively small number of parameters for specific challenges. In our paper, we design different fusion schemes according to the characteristics of each challenge, which have simple and basic structures. By decoupling the extracted features for different challenges, each fusion scheme only needs to learn a few parameters for a specific challenge and has less dependence on data.

	\subsection{RGBD Salient Object Detection}
	RGBD SOD methods typically use depth maps as supplementary information to enhance RGB modality. 
	Several deep learning-based methods are proposed to explore the sufficient fusion of cross-modal complementarity~\cite{qu2017rgbd,zhou2021specificity,chen2018progressively,fu2020jl,liu2020learning}. For example, Qu et al.~\cite{qu2017rgbd} design handcrafted features from input images and feed them into the convolutional neural network stream for saliency guidance. Chen et al.~\cite{chen2018progressively} design a three-stream network to enhance the representation of multi-modal features. Fu et al.~\cite{fu2020jl} fuse multi-modal features through a Siamese network to explore representative saliency features in the feature extraction phase. Liu et al.~\cite{liu2020learning} design a selective self-mutual attention that selectively supplements complementary information based on the information of the other modality. SPNet~\cite{zhou2021specificity} captures specific and shared features through a modality-specific branch and a modality-sharing branch, respectively. Zhang et al.~\cite{zhang2021uncertainty} achieve probabilistic cross-modal fusion through a generative architecture. Some methods also focus on addressing the challenge of depth ambiguity~\cite{zhao2019contrast,piao2019depth,fan2020rethinking,li2021hierarchical,Ji_2021_DCF,ji2022dmra}. CPFP~\cite{zhao2019contrast} designs a specific network to improve depth information, which is then fused with RGB features to enhance the contrast between foreground and background. Fan et al.~\cite{fan2020rethinking} publish a large-scale dataset of human activity scenarios and design a three-stream module to filter unreliable depth maps. DCF~\cite{Ji_2021_DCF} mitigates noise in low-quality depth maps through a calibration strategy. DMRA~\cite{ji2022dmra} utilizes RGB complementary cues to improve depth information. Recently, transformer-based methods~\cite{liu2021visual,zhang2022learning,pang2023caver,tang2022hrtransnet,liu2021swinnet} have demonstrated the superior performance due to their global modeling capability. VST~\cite{liu2021visual} introduces the transformer into the salient object detection field for the first time, which models the global dependency of objects. The recent transformer-based method CAVER~\cite{pang2023caver} designs a novel attention mechanism to integrate multi-scale and multi-modal features and propagate the global context. 
	
	The above methods exploit the complementary information between the modalities through different fusion strategies and obtain performance gains. However, they can still fall into performance bottlenecks when facing some challenging scenarios, such as depth ambiguity or thermal crossover, low illumination, image clutter, etc. Although some methods~\cite{pang2023caver,Ji_2021_DCF,fan2020rethinking} can solve specific challenges, they also fail to address multiple challenges simultaneously. To deal with the major challenges in multi-modal SOD, we propose the adaptive fusion bank, which integrates multiple fusion schemes tailored for different challenges and adaptively selects appropriate fusion schemes according to the input data.

\section{The Proposed Method}
\label{sec:method}
\begin{figure*}[t]
	\centering
	\includegraphics[width=1\linewidth]{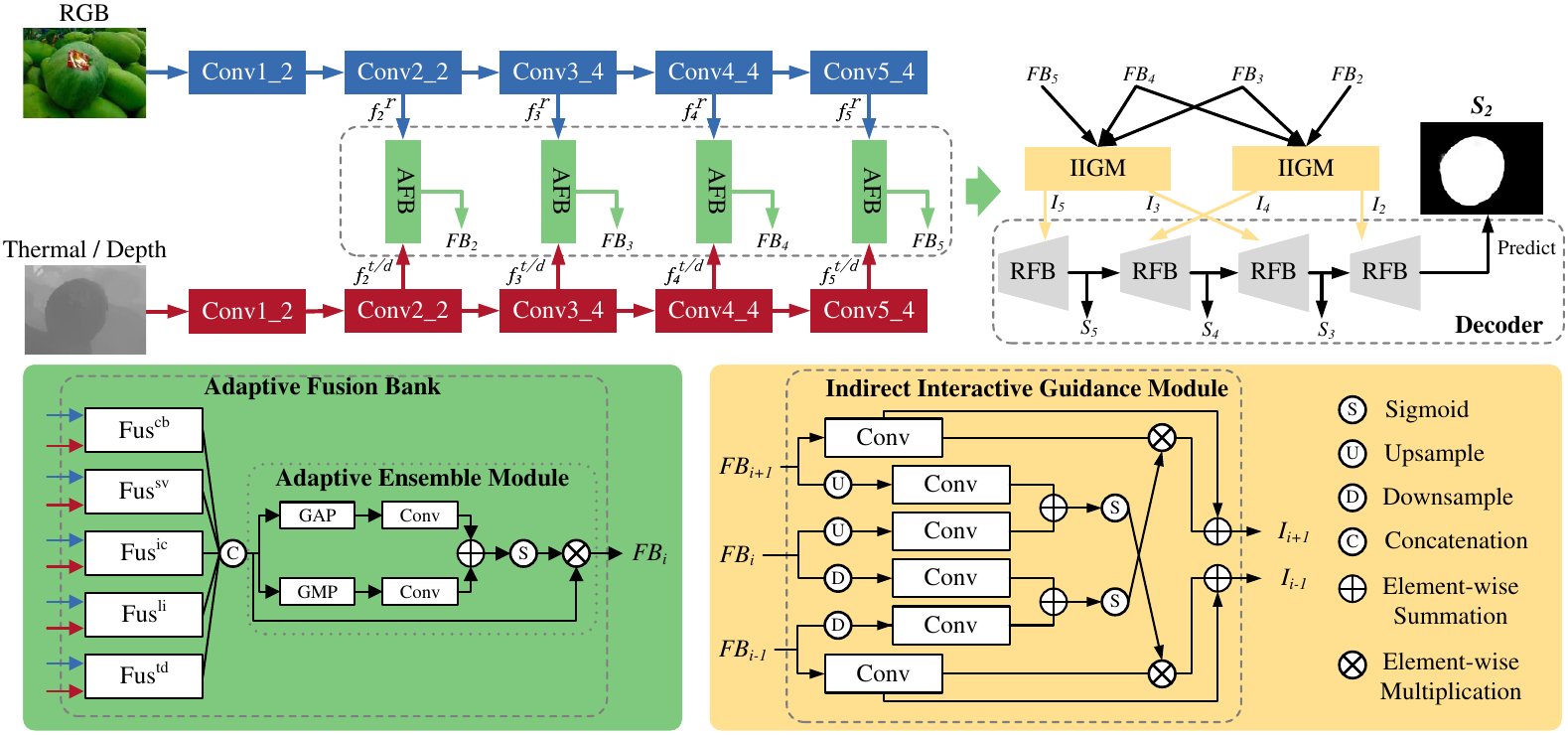}
	\caption{Overview of our proposed learning adaptive fusion bank (LAFB). 
		We first send the extracted multi-modal 
		features (i.e., 
		$f_i^r$ and $f_i^{t/d}$) into the adaptive 
		fusion bank (AFB), which contains five specific fusion schemes (i.e., $Fu{s^{cb}}$, $Fu{s^{sv}}$, $Fu{s^{ic}}$, $Fu{s^{li}}$ and $Fu{s^{td}}$) to generate multi-level features (i.e., ${FB_i}$) for corresponding challenges. Then, the multi-level features are fed into the 
			indirect interactive guidance module (IIGM) to integrate 
			high-level and low-level features smoothly. After that, the generated features (i.e., $I_i$) are fed into the RFB module to increase the receptive field of features. Finally, multi-level saliency maps (i.e., ${S_i}$) are inferred in a top-down manner in the decoder, and ${S_2}$ is taken as the final saliency map.} 
	\label{fig::FDNet}
\end{figure*}

In this section, we present an overview of our proposed learning adaptive fusion bank (LAFB) and elaborate on its key components, including the adaptive fusion bank and the indirect interactive guidance module. We also describe the loss function used in our method.

\subsection{Overview}

As shown in Fig.~\ref{fig::FDNet}, our LAFB model is based on an encoder-decoder framework. First, the multi-modal images are separately fed into the encoder, which is built on the Res2Net-50~\cite{gao2019res2net} network. The encoder extracts hierarchical multi-modal features, denoted as $f_i^r$ and $f_i^{t/d}$, where $r$ represents the RGB modality, $t/d$ represents the thermal or depth modality, and $i \in \{ 1,2,...,5\} $ denotes different layers. To reduce computation, we discard the lowest feature maps, i.e., $f_1^r$ and $f_1^{t/d}$. Next, in the decoder stage, we use top-down inference to integrate the hierarchical features and predict saliency maps.

Adaptive fusion bank (AFB) and indirect interactive guidance module (IIGM) are the two key components of our LAFB. AFB is designed to 
leverage the complementary benefits of multiple fusion schemes for addressing different challenges. As shown in Fig.~\ref{fig::FDNet}, each bank consists of five different fusion schemes and an adaptive ensemble module. The extracted features are sent into AFB and decoupled by different fusion schemes for different challenges. Subsequently, the adaptive ensemble module aggregates the decoupled features to address current challenges, generating the multi-level features ${FB_i}$. Then, the IIGM takes three adjacent features (i.e., ${FB_{i-1}}$, ${FB_i}$ and ${FB_{i+1}}$, $i \in \{ 3,4\} $) as inputs to effectively integrate high-level semantic information and low-level detailed features. The integrated features ${I_i}$ are then sent into the decoder to generate the final saliency map $S_2$ in a top-down manner.

\subsection{Adaptive Fusion Bank}

In the multi-modal salient object detection (MSOD) task, the design of fusion schemes can affect the quality of the fused features and the model performance.
Existing MSOD methods~\cite{zhang2019rgb,tu2020rgbt,huo2021efficient,qu2017rgbd,zhang2021uncertainty} focus on designing complex fusion strategies to exploit multi-modal complementarity, which achieve performance improvements. However, they also fall into performance bottlenecks when faced with challenging scenarios, such as those with depth ambiguity, low illumination, image clutter, etc. To solve this problem, some recent methods~\cite{wang2021cgfnet,liao2022cross,zhou2021ecffnet,liao2022cross} design fusion strategies to cope with certain challenges. Nonetheless, these fusion strategies are tailored for specific challenges, and they hardly generate rich salient cues in other common challenging scenes.


To address multiple major challenges simultaneously, we propose to design a fusion bank comprising different fusion schemes that can adaptively select appropriate schemes for different challenges in the input data.
There are two important issues that need to be addressed. First, how to design these fusion schemes so that can effectively handle different challenges. Second, how to select appropriate fusion schemes for the different challenges of multi-modal inputs. Therefore, we design the adaptive fusion bank to solve these two issues, as depicted in Fig.~\ref{fig::FDNet}.

\begin{figure}[t]
	\centering
	\includegraphics[width=\linewidth]{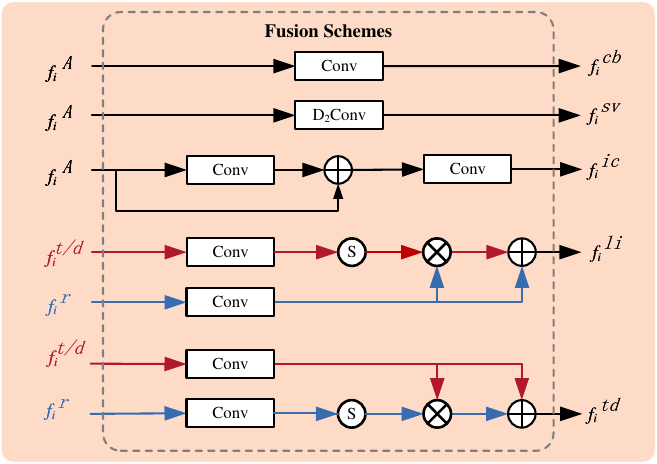}
	\caption{Architectures of the fusion schemes in the adaptive fusion 
		bank. $f_i^A$ represents the concatenation of multi-modal 
			features (i.e., $f_i^{t/d}$ and $f_i^r$) extracted by the backbone. $f_i^{cb}$, $f_i^{sv}$, $f_i^{ic}$, $f_i^{li}$ and $f_i^{td}$ are generated by the corresponding fusion schemes for different challenges.}
	\label{fig::AFB}
\end{figure}

{\bf Fusion schemes:} Building upon the work~\cite{tu2020rgbt}, which has described the challenges of RGBT SOD in detail, we identify the main challenges in MSOD as follows: modality-shared challenges, including center bias (CB), scale variation (SV), and image clutter (IC), and modality-specific challenges, including low illumination (LI), thermal crossover or depth ambiguity (TD). Considering the different characteristics of these main challenges, we design the corresponding five basic fusion schemes to solve the first problem mentioned above, as shown in Fig.~\ref{fig::AFB}. The fusion schemes are designed to be simple in structure. There are several reasons for this. First, considering the overall computation complexity of the adaptive fusion bank that is embedded into each encoder layer, we design each fusion scheme with simple but effective structure. Second, the limited amount of multi-modal data makes it difficult to learn a large number of parameters for each challenge using complex fusion schemes, while a few parameters in simple fusion schemes are easier to be learned. Third, these fusion schemes share the same backbone, which extracts rich and common features. Hence, each fusion scheme only requires a simple tailored design to decouple effective features for the specific challenge.

Center bias is a relatively simple challenge, objects with center-bias are usually close to the image boundary and mainly need to be located. Although complex and deep fusion structures have a strong positioning ability, they also have high computational complexity. Considering the overall computational complexity of AFB, we adopt a simple and effective design for $Fu{s^{cb}}$.
Since the backbone has already extracted rich features, the center bias fusion scheme only needs to decouple the location information of the object from it.
Therefore, we use a convolutional block with a 3×3 kernel size to extract the extracted features at each location by sliding, so that the features of objects can be captured as much as possible. In addition, center bias is a common challenge that usually accompanied with other challenges~\cite{tu2020rgbt}. By combining with other fusion schemes, $Fu{s^{cb}}$ is able to more accurately locate salient objects with the help of global features generated by other complex fusion schemes. This fusion scheme, denoted as 
$Fu{s^{cb}}$, is designed to address center bias and can be formulated as follows:
\begin{equation}\label{Eq::CB}
	\begin{split}
		&f_i^{cb} = Conv(f_i^A),
	\end{split}
\end{equation}
where $f_i^A = [f_i^r,f_i^{t/d}] $ represents the concatenation of multi-modal features extracted by the backbone, and $Conv(.)$ means a convolution layer. $f_i^{cb}$ is the output of the $i_{th}$ multi-modal feature $f_i^A$ through the fusion scheme $Fu{s^{cb}}$.

In the case of the scale variation challenge, the size of salient objects are variable and the number of salient objects can be single or multiple. Therefore, a wide receptive field is necessary to fully capture the scale-varying objects. The original convolutional block focuses on local areas and is difficult to cover the large or multiple objects effectively. To address this, we utilize a dilated convolution, which is specialized in increasing the receptive field and beneficial for capturing the large or multiple objects. Specifically, we use a convolutional layer with a 3 × 3 kernel size and a dilation rate of 2 for the $Fu{s^{sv}}$ fusion scheme. This process can be formulated as follows:
\begin{equation}\label{Eq::SV}
	\begin{split}
		&f_i^{sv} = {D_2}Conv(f_i^A),
	\end{split}
\end{equation}
where $f_i^{sv}$ denotes the features generated by the fusion scheme $Fu{s^{sv}}$, ${D_2}Conv(.)$ is the dilated convolution with dilation rate 2.

Image clutter is a relatively difficult challenge, caused by low contrast 
between foreground and background, complex background, etc. As shown in the fifth row of Fig.~\ref{fig::branch}, under the image clutter challenge, too much noise exists in the background, which prevents the network from detecting salient objects accurately. To this end, we first localize the salient regions in an image. Specifically, we use two consecutive convolutional blocks to obtain features with a larger field of view containing global and semantic information that can help identify regions of the image. Second, we finely separate the salient regions from the background. Specifically, we introduce the initial input features through a residual connection. Due to the relatively small receptive field, the initial features have rich spatial details, such as textures, boundaries, etc., which can make salient and non-salient areas more distinguishable. In this way, salient objects in the image clutter challenge scene can be accurately recognized and separated out.
We obtain the output feature $f_i^{ic}$ of this fusion scheme $Fu{s^{ic}}$ as follows:
\begin{equation}\label{Eq::IC}
	\begin{split}
		&f_i^{ic} = Conv(Conv(f_i^A) + f_i^A).
	\end{split}
\end{equation} 
\begin{figure}[t]
	\centering
	\includegraphics[width=1\columnwidth]{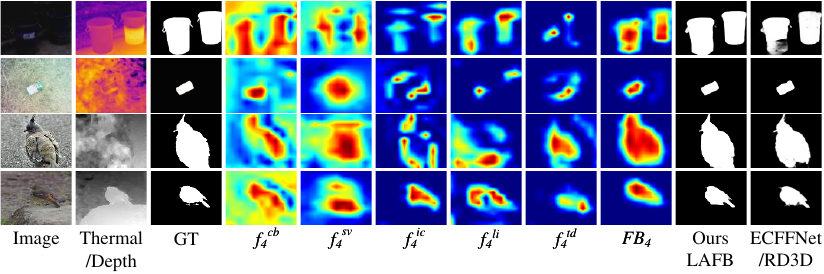}
	\caption{Visualization of feature maps generated by different fusion 
		schemes and adaptive ensemble module in the ${5^{th}}$ adaptive 
		fusion bank. 
		From the ${1^{st}}$ row to the ${4^{th}}$ row, the challenge is low light, thermal 
		crossover, depth ambiguity and image clutter, respectively. The results of the saliency maps indicate that our method is able to deal with multiple challenges simultaneously, while the comparison methods (i.e., ECFFNet~\cite{zhou2021ecffnet} and EBFSP~\cite{huang2021employing}) fails on some challenges, such as the first and the third rows.}
	\label{fig::visual_cp}
\end{figure}
\begin{figure}[t]
	\centering
	\includegraphics[width=\columnwidth]{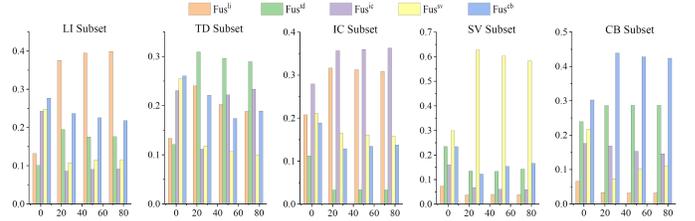}
	\caption{Weights assigned to different fusion schemes when training on 
		different challenge data, which are all obtained based on the challenge annotation of the training set of VT5000. The x-axis and y-axis represent epochs and 
		weights, respectively. The values of columns in different colors indicate the weight assigned to different fusion schemes during training.}
	\label{fig::weight_RGBT}
\end{figure}

Low illumination is a challenge specific to the RGB modality. In this case, it is difficult to capture a clear object appearance from RGB images, while thermal or depth images tend to contain more sufficient object information. Therefore, it is challenging for RGB modality to cope with this type of challenge due to feature degradation. To address this challenge, we introduce the features of the thermal or the depth modality to 
compensate for the lack of information. In detail, we first smooth the 
features of the thermal or the depth modality with a 3 × 3 convolutional block to reduce noise in them. Since thermal or depth modality captures objects more explicitly than RGB modality in the low illumination scenario, we use a Sigmoid activation function $\sigma $ to obtain the weight distribution ${W_{td}}$ of the thermal or the depth features, highlighting the object's location. Then, we perform the multiplication of the weight ${W_{td}}$ and RGB features to guide RGB features to focus on the object region. Considering the information loss caused by multiplication, we maintain RGB information by summing the original RGB features with the guided RGB features.
The process of the fusion scheme 
$Fu{s^{li}}$ 
can be described as follows:
\begin{equation}\label{Eq::LI_weight}
	\begin{split}
		&{W_{td}} = \sigma (Conv(f_i^{t/d})),
	\end{split}
\end{equation} 
\begin{equation}\label{Eq::LI}
	\begin{split}
		&f_i^{li} = {W_{td}} \odot f_i^r + f_i^r,
	\end{split}
\end{equation} 
where $\odot $ represents an element-wise multiplication.

Thermal crossover or depth ambiguity is another modality-specific challenge. In this case, the object features of thermal or depth images are degenerated, while RGB images tend to have more complete object information. To deal with this challenge, we use the features of the RGB modality as an information 
supplement. Hence, we design a fusion scheme $Fu{s^{td}}$ similar to the fusion 
scheme used for the low illumination challenge. This process can be 
formulated as:
\begin{equation}\label{Eq::TC_weight}
	\begin{split}
		&{W_r} = \sigma (Conv(f_i^r)),
	\end{split}
\end{equation} 
\begin{equation}\label{Eq::TC}
	\begin{split}
		&f_i^{td} = {W_r} \odot f_i^{t/d} + f_i^{t/d}.
	\end{split}
\end{equation}
Note that although the modality-specific fusion schemes are similar to the spatial attention mechanism~\cite{woo2018cbam} and the cross-modal fusion of the method~\cite{zhao2019contrast}, their impact is completely different. The spatial attention mechanism and the cross-modal fusion of the method~\cite{zhao2019contrast} both refine feature representations in general, without a specific objective. Instead, our modality-specific fusion schemes are tailored for low illumination, thermal crossover, or depth ambiguity challenges. By being combined with other fusion schemes and adaptively assigned training weights based on the challenges in the input data through AEM, the modality-specific fusion schemes can be effectively trained and further decouple the extracted features to deal with the modality-specific challenges.

{\bf Adaptive ensemble module:} 
Although each fusion scheme can address a specific challenge, the 
	input data is dominated by different challenges, which means that each input data may require a different combination of fusion schemes. Therefore, it is necessary to select appropriate fusion schemes according to the challenges present in the input data. To address this issue, we propose an adaptive ensemble module (AEM) to select the 
fusion schemes that effectively address challenges in 
the input data. AEM integrates these fusion schemes by learning the 
weights of features generated by them. As illustrated in 
Fig.~\ref{fig::FDNet}, AEM is designed inspired by channel 
attention~\cite{woo2018cbam}, which is usually used to refine independent 
features. Unlike channel attention, our AEM is responsible for adaptively integrating 
features that can effectively deal with different challenges.
In this way, the fusion schemes designed for current challenges can be 
selected effectively.

Specifically, we first concatenate all the fused features of the fusion 
schemes in 
the channel dimension, which is denoted as $f_i^{C}$. Then, a global 
average 
pooling layer ($GAP$) and a global max pooling layer ($GMP$) followed by 
one 
1 × 1 convolutional layer are separately adopted, which compress the spatial 
dimension of the 
concatenated feature from two aspects. After that, we perform pixel-wise 
addition operation on the two compressed 
features and then utilize a Sigmoid function to 
learn the channel weight vector $V \in {\mathbb{R}^{c \times 1 \times 
		1}}$, in which the $c$ represents the number of channels of the 
feature. The formulation of this 
procedure can be expressed as:
\begin{equation}\label{Eq::ca0}
	\begin{split}
		&f_i^C = [f_i^{cb},f_i^{sv},f_i^{ic},f_i^{li},f_i^{td},],
	\end{split}
\end{equation} 
\begin{equation}\label{Eq::ca1}
	\begin{split}
		&V = \sigma (Conv(GAP(f_i^C)) + Conv(GMP(f_i^C))),
	\end{split}
\end{equation} 
where $[,.,]$ represents the channel-wise concatenation. Then the features 
of each fusion scheme is weighted by the vector $V$, as:
\begin{equation}\label{Eq::ca2}
	\begin{split}
		&F{B_i} = V \odot f_i^C,
	\end{split}
\end{equation} 
where $F{B_i}$ is the output feature of the $i_{th}$ layer fusion bank. At 
this 
point, the second problem mentioned above has also been solved.

In Fig.~\ref{fig::visual_cp}, the visualization results show that the extracted 
multi-modal features decoupled by different fusion schemes are specialized in 
handling different challenges. For each multi-modal input, the fusion schemes for the challenges in the input data can decouple the extracted features with more saliency cues. Besides, the 
feature maps generated by AEM always have the highest response to the 
objects. This 
demonstrates that the fusion schemes are effective in addressing 
different challenges and the AEM can make them complementary to 
achieve 
adaptive integration.

To further illustrate that the proposed fusion schemes in AFB can be trained specifically,
we train the network separately with different challenge datasets annotated
in the training set of VT5000~\cite{tu2020rgbt}. During 
training, we 
obtain the weight assigned to each fusion scheme, which is 
separated from the weight vector $V$. The weight changes 
of different fusion schemes on different challenge datasets are 
shown in Fig.~\ref{fig::weight_RGBT}. Note that each subset of challenge data is derived from the training set of the VT5000~\cite{tu2020rgbt} and is obtained according to its challenge annotation. More details regarding the challenge annotation can be referred to~\cite{tu2020rgbt}. 
It can be seen that AFB assigns larger weights to the fusion schemes tailored to the current input challenges, since they can generate features with more salient cues. In this way, for different input data, the corresponding fusion schemes can be trained with bigger gradients. Furthermore, with the iteration of training, each fusion scheme can learn corresponding parameters to deal with corresponding challenges more effectively, and AEM will also assign a greater weight to the fusion scheme that meets the current challenges, as shown in Fig. 5. Therefore, each fusion scheme can be fully trained by the corresponding challenge data to deal with their respective challenges.

\subsection{Indirect Interactive Guidance Module}
After the adaptive fusion bank integrates the multi-modal features, hierarchical features $F{B_i}$ are obtained. Low-level detailed features contain rich detailed information, such as texture, color, boundary, etc., which can make the foreground and background more distinguishable. High-level features provide sufficient semantic information that facilitates category recognition and the localization of salient regions from background. By integrating multi-level features, richer features can be obtained for accurate segmentation in complex scenes, such the hollow objects. Existing methods~\cite{wang2021cgfnet,Ji_2021_DCF,zhou2021ccafnet,zhou2021specificity,liu2021visual} often use top-down fusion to integrate features layer by layer, which limits feature variety and prevents full fusion of high-level and low-level features. Some other methods~\cite{Sun2021DeepRS,zhang2020feature} directly integrate multi-level features to generate richer features for each layer, but they fail to fully exploit the unique characteristics of high-level and low-level features. Furthermore, direct integration of multi-level features may compromise the original information contained in these features. 

To address the above problem, we propose an indirect interactive guidance module that exploits the unique characteristics of high-level and low-level features to achieve mutual guidance of multi-level features. Besides, by introducing an intermediate feature,  high-level and low-level features can be integrated indirectly, so as to preserve the origin information in them as much as possible. 
%
As shown in Fig.~\ref{fig::FDNet}, we use three 
successive multi-modal features ($F{B_{i-1}}$, $F{B_i}$ and 
$F{B_{i+1}}$, $i \in \{ 3,4\} $) as inputs, which contain both high-level semantic 
features ($F{B_{4}}$ and $F{B_{5}}$) and low-level detail features ($F{B_{2}}$ and 
$F{B_{3}}$). Then, we take the intermediate $F{B_i}$ as a 
springboard to fuse the high-level feature $F{B_{i+1}}$ with 
the low-level feature $F{B_{i-1}}$ interactively. With this springboard, 
high-level features and low-level features can be smoothly associated without 
intermediate information gaps. In detail, on the one hand, 
we first up-sample 
$F{B_i}$ 
and 
$F{B_{i+1}}$ to the 
size of $F{B_{i-1}}$, and send them into the 
convolutional layer with a 3 × 
3 
kernel size separately. Then we fuse them by a pixel-wise addition and apply a 
Sigmoid function to learn the high-level weight $W_{high}$, which is used to 
guide the low-level feature $F{B_{i-1}}$. The whole 
process can be 
formulated as follows:
\begin{equation}\label{Eq::mask1}
	\begin{split}
		&{W_{high}} = \sigma (Conv(Us(F{B_i})) + Conv(Us(F{B_{i+1}}))).
	\end{split}
\end{equation} 
$Us(.)$ indicates the upsample operation with bilinear interpolation. On 
the 
other 
hand, we 
down-sample $F{B_i}$ and $F{B_{i-1}}$ to the 
size of $F{B_{i+1}}$, then do same 
operations as 
above 
to generate the weight to guide 
$F{B_{i+1}}$, as follows:
\begin{equation}\label{Eq::mask2}
	\begin{split}
		&{W_{low}} = \sigma (Conv(Ds(F{B_{i - 1}})) + Conv(Ds(F{B_i}))),
	\end{split}
\end{equation} 
where $W_{low}$ is the low-level weight and $Ds(.)$ indicates the 
down-sample 
operation with bilinear interpolation. Then, we perform element-wise 
multiplication of $W_{low}$ with $F{B_(i+1)}$ and $W_{high}$ with $F{B_{i-1}}$ 
respectively to form the guidance.
To retain the details, we add the original features $F{B_{i+1}}$ and 
$F{B_{i-1}}$, separately. The whole process can be 
formulated as follows:
\begin{equation}\label{Eq::IIGM1}
	\begin{split}
		&{I_{i + 1}} = F{B_{i+1}} + F{B_{i+1}} \odot {W_{low}},\\
		&{I_{i - 1}} = F{B_{i-1}} + F{B_{i-1}} \odot {W_{high}},
	\end{split}
\end{equation} 
where ${I_{i+1}}$ and ${I_{i-1}}$ are the output features of the indirect 
interactive 
guidance module.

In the decoder, the integrated features ${I_{i}}$ are fused in a top-down 
manner, as in 
U-Net~\cite{ronneberger2015u}. 
Besides, following~\cite{zhou2021specificity}, we embed the 
receptive field block (RFB)~\cite{wu2019cascaded} into 
the process to increase the receptive field of features. Then the final 
saliency map $S_2$ is predicted from the lowest fused feature by a 1 × 1 
convolutional layer and an upsample operation.

\subsection{Loss Function}
Similar to previous works~\cite{zhou2021specificity,tu2021multi}, we employ the binary 
cross-entropy loss to calculate the distance between the 
final predicted saliency map and the ground truth. To allow our model to 
learn discriminative information, we also use the binary cross-entropy loss 
on 
the 
remaining features of the decoder, i.e.,
\begin{equation}\label{Eq::bce2}
	\begin{split}
		&{\ell _i} = {\ell _{bce}}({S_i},G),i = 2,3,4,5.
	\end{split}
\end{equation} 
Where ${\ell _{bce}}$ represents the binary cross-entropy loss, $G$ is the 
ground truth, $S_i$ consists the final saliency map $S_2$ and the intermediate 
saliency 
maps 
$\left\{ {{S_i}} \right\}_{i = 3}^5$ produced from the remaining features 
of 
the decoder.

Following~\cite{tu2021multi} and~\cite{wu2021mobilesal}, we also use the 
smoothness 
loss~\cite{godard2017unsupervised} $\ell 
_s$ 
and dice loss~\cite{milletari2016v} $\ell _d$ for the final saliency map 
$S_2$ 
to keep the boundary 
clear and the region consistent.
Therefore, the 
total loss function can be expressed as:
\begin{equation}\label{Eq::total loss}
	\begin{split}
		&{\ell _{total}} = \sum\limits_{i = 2}^5 {{\lambda _i}{\ell 
				_i}}  + {\ell _s} + {\ell _d},
	\end{split}
\end{equation}
where ${\lambda _i}$ denote the weights of different 
losses, and we set 
the values of ${\lambda _2}$, ${\lambda _3}$, ${\lambda _4}$ and 
${\lambda 
	_5}$ to 1.0, 0.8, 0.6, 0.5 respectively. The 
ablation study and specific analysis are presented in 
Sec.~\ref{ablation}.
\begin{table*}[htbp]
	\centering
	\caption{Quantitative Comparison of E-Measure (${E_\xi }$), Weighed 
		F-Measure 
		(${F_\beta ^\omega}$), F-Measure 
		(${F_\beta }$), Mean Absolute Error 
		($MAE$) and Frame-Per-Second (FPS) with 17 
		Different RGBD Methods on seven 
		Testing Datasets, Including SIP~\cite{fan2020rethinking}, 
		DUT-RGBD~\cite{piao2019depth} (i.e., DUT), SSD~\cite{zhu2017three}, STERE~\cite{niu2012leveraging}, ReDWeb-S~\cite{liu2021learning}, NLPR~\cite{peng2014rgbd}, and NJUD~\cite{ju2014depth}. The Best, Second 
		Best and Third Best 
		Results Are 
		Marked with \textcolor[RGB]{237,31,36}{Red} and \textcolor[RGB]{105,189,69}{Green}, Respectively. ‘-’ Indicates the Code or Result Is Not Available.}
	\setlength{\tabcolsep}{4pt}
	\renewcommand\arraystretch{1.5}
	\resizebox{1\textwidth}{!}{\Huge
		\begin{tabular}{c|c|cccc|cccc|cccc|cccc|cccc|cccc|cccc|c|c|c}
			\hline
			\multirow{2}[4]{*}{Methods} & \multirow{2}[4]{*}{Backbone} & \multicolumn{4}{c|}{SIP~\cite{fan2020rethinking}}      & \multicolumn{4}{c|}{DUT~\cite{piao2019depth}}      & \multicolumn{4}{c|}{SSD~\cite{zhu2017three}}      & \multicolumn{4}{c|}{STERE~\cite{niu2012leveraging}}    & \multicolumn{4}{c|}{ReDWeb-S~\cite{liu2021learning}} & \multicolumn{4}{c|}{NLPR~\cite{peng2014rgbd}}     & \multicolumn{4}{c|}{NJUD~\cite{ju2014depth}}     & \multirow{2}[4]{*}{FPS $\uparrow$} & \multirow{2}[4]{*}{Param $\downarrow$} & \multirow{2}[4]{*}{GFLOPs $\downarrow$} \bigstrut\\
			\cline{3-30}          &       & ~${E_\xi}$ \hfill$\uparrow$~     & ${F_\beta ^\omega}$ \hfill$\uparrow$~    & ${F_\beta}$ \hfill$\uparrow$     & $MAE$ \hfill$\downarrow$   & ~${E_\xi}$ \hfill$\uparrow$~     & ${F_\beta ^\omega}$ \hfill$\uparrow$~    & ${F_\beta}$ \hfill$\uparrow$     & $MAE$ \hfill$\downarrow$   & ~${E_\xi}$ \hfill$\uparrow$~     & ${F_\beta ^\omega}$ \hfill$\uparrow$~    & ${F_\beta}$ \hfill$\uparrow$     & $MAE$ \hfill$\downarrow$   & ~${E_\xi}$ \hfill$\uparrow$~     & ${F_\beta ^\omega}$ \hfill$\uparrow$~    & ${F_\beta}$ \hfill$\uparrow$     & $MAE$ \hfill$\downarrow$   & ~${E_\xi}$ \hfill$\uparrow$~     & ${F_\beta ^\omega}$ \hfill$\uparrow$~    & ${F_\beta}$ \hfill$\uparrow$     & $MAE$ \hfill$\downarrow$   & ~${E_\xi}$ \hfill$\uparrow$~     & ${F_\beta ^\omega}$ \hfill$\uparrow$~    & ${F_\beta}$ \hfill$\uparrow$     & $MAE$ \hfill$\downarrow$   & ~${E_\xi}$ \hfill$\uparrow$~     & ${F_\beta ^\omega}$ \hfill$\uparrow$~    & ${F_\beta}$ \hfill$\uparrow$     & $MAE$ \hfill$\downarrow$   &       &       &  \bigstrut\\
			\hline
			HDFNet$_{20}$~\cite{HDFNet-ECCV2020} & VGG19 & 0.920  & 0.848  & 0.862  & 0.048  & 0.937  & 0.873  & 0.887  & 0.040  & 0.902  & 0.823  & 0.831  & 0.047  & 0.924  & 0.863  & 0.860  & 0.041  & 0.760  & 0.623  & 0.704  & 0.133  & 0.949  & 0.873  & 0.874  & 0.028  & 0.911  & 0.879  & 0.856  & 0.039  & 77    & 164.19  & 111.02  \bigstrut[t]\\
			CoNet$_{20}$~\cite{ji2020accurate} & ResNet101 & 0.909  & 0.814  & 0.842  & 0.063  & 0.948  & 0.896  & 0.909  & 0.033  & 0.896  & 0.792  & 0.806  & 0.059  & 0.928  & 0.874  & 0.885  & \textcolor[RGB]{105,189,69}{0.037} & 0.762  & 0.618  & 0.688  & 0.147  & 0.934  & 0.850  & 0.848  & 0.031  & 0.912  & 0.856  & 0.873  & 0.046  & 154   & 162.13  & 13.05  \\
			CCAFNet$_{21}$~\cite{zhou2021ccafnet} & ResNet50 & 0.915  & 0.839  & 0.864  & 0.054  & 0.940  & 0.884  & 0.903  & 0.037  & 0.915  & 0.839  & \textcolor[RGB]{105,189,69}{0.864} & 0.054  & 0.921  & 0.853  & 0.869  & 0.044  & 0.687  & 0.515  & 0.614  & 0.173  & 0.951  & 0.883  & 0.880  & 0.026  & 0.920  & 0.883  & 0.896  & 0.037  & 88    & 159.45  & 189.47  \\
			CDNet$_{21}$~\cite{jin2021cdnet} & VGG16 & 0.913  & 0.839  & 0.870  & 0.056  & 0.936  & 0.878  & 0.901  & 0.039  & 0.849  & 0.706  & 0.742  & 0.073  & 0.929  & 0.871  & 0.884  & 0.039  & 0.730  & 0.588  & 0.678  & 0.146  & 0.951  & 0.886  & 0.879  & 0.025  & 0.903  & 0.828  & 0.856  & 0.054  & 86    & 123.65  & 72.07  \\
			HAINet$_{21}$~\cite{li2021hierarchical} & VGG16 & 0.924  & 0.860  & 0.883  & 0.049  & 0.937  & 0.887  & 0.905  & 0.038  & 0.843  & 0.682  & 0.735  & 0.101  & \textcolor[RGB]{105,189,69}{0.930} & 0.877  & 0.888  & 0.038  & \textcolor[RGB]{237,31,36}{0.766} & 0.654  & 0.713  & 0.132  & 0.951  & 0.884  & 0.884  & 0.025  & 0.917  & 0.882  & 0.895  & 0.039  & 11    & 228.20  & 181.62  \\
			DFM$_{21}$~\cite{zhang2021depth} & MobileNetV2 & 0.919  & 0.844  & 0.871  & 0.051  & 0.898  & 0.795  & 0.830  & 0.062  & 0.871  & 0.733  & 0.755  & 0.076  & 0.912  & 0.850  & 0.858  & 0.045  & 0.753  & 0.610  & 0.690  & 0.136  & 0.945  & 0.876  & 0.868  & 0.026  & 0.913  & 0.868  & 0.885  & 0.042  & 252   & \textcolor[RGB]{237,31,36}{8.45} & 5.43  \\
			SPNet$_{21}$~\cite{zhou2021specificity} & Res2Net50 & \textcolor[RGB]{105,189,69}{0.930} & 0.873  & \textcolor[RGB]{105,189,69}{0.891} & \textcolor[RGB]{105,189,69}{0.043} & 0.876  & 0.747  & 0.843  & 0.085  & 0.910  & 0.831  & 0.852  & \textcolor[RGB]{105,189,69}{0.044} & \textcolor[RGB]{105,189,69}{0.930} & 0.879  & 0.886  & \textcolor[RGB]{105,189,69}{0.037} & 0.759  & 0.637  & 0.712  & 0.129  & 0.957  & 0.899  & \textcolor[RGB]{105,189,69}{0.898} & \textcolor[RGB]{237,31,36}{0.021} & \textcolor[RGB]{237,31,36}{0.931} & \textcolor[RGB]{105,189,69}{0.909} & 0.915  & \textcolor[RGB]{105,189,69}{0.029} & 50    & 573.39  & 68.10  \\
			RD3D$_{21}$~\cite{chen2021rgb} & I3DResNet50 & 0.919  & 0.852  & 0.873  & 0.049  & 0.949  & 0.913  & 0.925  & 0.031  & 0.905  & 0.794  & 0.812  & 0.052  & 0.926  & 0.877  & 0.885  & 0.038  & 0.701  & 0.487  & 0.595  & 0.177  & 0.957  & 0.894  & 0.890  & \textcolor[RGB]{105,189,69}{0.022} & 0.918  & 0.890  & 0.900  & 0.037  & 94    & 110.30  & 43.51  \\
			DSA2F$_{21}$~\cite{Sun2021DeepRS} & VGG19 & 0.908  & 0.838  & 0.867  & 0.057  & 0.950  & 0.914  & \textcolor[RGB]{105,189,69}{0.926} & 0.030  & 0.904  & 0.836  & 0.852  & 0.047  & 0.928  & 0.877  & \textcolor[RGB]{105,189,69}{0.895} & 0.038  & -     & -     & -     & -     & 0.950  & 0.889  & 0.897  & 0.024  & 0.923  & 0.889  & 0.901  & 0.039  & -     & -     & - \\
			DCF$_{21}$~\cite{Ji_2021_DCF} & ResNet50 & 0.920  & 0.850  & 0.877  & 0.051  & 0.952  & 0.913  & 0.925  & 0.030  & 0.898  & 0.800  & 0.829  & 0.053  & \textcolor[RGB]{237,31,36}{0.931} & 0.880  & 0.890  & \textcolor[RGB]{105,189,69}{0.037} & 0.755  & 0.632  & 0.710  & 0.135  & 0.956  & 0.892  & 0.893  & 0.023  & 0.922  & 0.884  & 0.897  & 0.038  & 57    & 111.51  & 59.38  \\
			MobileSal$_{21}$~\cite{wu2021mobilesal} & MobileNetV2 & 0.914  & 0.837  & 0.860  & 0.054  & 0.936  & 0.869  & 0.912  & 0.044  & 0.898  & 0.804  & 0.815  & 0.052  & 0.916  & 0.865  & 0.848  & 0.041  & 0.671  & 0.455  & 0.608  & 0.186  & 0.950  & 0.878  & 0.872  & 0.025  & 0.914  & 0.874  & 0.894  & 0.040  & \textcolor[RGB]{105,189,69}{268} & 24.97  & \textcolor[RGB]{105,189,69}{1.96} \\
			SSL$_{22}$~\cite{zhao2022self} & VGG16 & 0.921  & 0.851  & 0.875  & 0.049  & 0.927  & 0.859  & 0.888  & 0.046  & 0.833  & 0.638  & 0.696  & 0.100  & 0.923  & 0.864  & 0.875  & 0.042  & -     & -     & -     & -     & 0.954  & 0.885  & 0.884  & 0.027  & 0.881  & 0.786  & 0.821  & 0.065  & 52    & 282.95  & 272.20  \\
			DIGRNet$_{22}$~\cite{cheng2022depth} & ResNet50 & 0.918  & 0.849  & 0.878  & 0.053  & 0.948  & 0.902  & 0.919  & 0.033  & 0.889  & 0.804  & 0.823  & 0.053  & 0.927  & 0.877  & 0.888  & 0.038  & -     & -     & -     & -     & 0.955  & 0.895  & 0.888  & 0.023  & \textcolor[RGB]{105,189,69}{0.928} & \textcolor[RGB]{105,189,69}{0.909} & \textcolor[RGB]{105,189,69}{0.916} & \textcolor[RGB]{237,31,36}{0.028} & 33    & 635.79  & 68.16  \\
			CIRNet$_{22}$~\cite{cong2022cir} & ResNet50 & 0.917  & 0.848  & 0.874  & 0.053  & 0.951  & 0.908  & \textcolor[RGB]{105,189,69}{0.926} & 0.031  & 0.898  & 0.791  & 0.821  & 0.054  & 0.921  & 0.836  & 0.875  & 0.049  & 0.725  & 0.519  & 0.630  & 0.171  & 0.955  & 0.889  & 0.883  & 0.023  & 0.922  & 0.881  & 0.896  & 0.040  & 93    & 393.50  & 42.60  \\
			MoADNet$_{22}$~\cite{jin2022moadnet} & MobileNetV3 & 0.908  & 0.828  & 0.850  & 0.058  & 0.949  & 0.911  & 0.923  & 0.031  & 0.894  & 0.801  & 0.824  & 0.057  & 0.914  & 0.861  & 0.868  & 0.042  & -     & -     & -     & -     & 0.945  & 0.875  & 0.874  & 0.027  & 0.909  & 0.881  & 0.892  & 0.041  & 167   & 19.19  & \textcolor[RGB]{237,31,36}{1.32} \\
			LSNet$_{23}$~\cite{zhou2023lsnet} & MobileNetV2 & 0.927  & 0.856  & 0.882  & 0.049  & 0.891  & 0.775  & 0.831  & 0.074  & 0.902  & 0.796  & 0.820  & 0.055  & 0.913  & 0.827  & 0.854  & 0.054  & 0.691  & 0.458  & 0.566  & 0.193  & 0.955  & 0.881  & 0.882  & 0.024  & 0.922  & 0.885  & 0.899  & 0.038  & \textcolor[RGB]{237,31,36}{316} & \textcolor[RGB]{105,189,69}{17.41} & 3.04  \\
			CAVER$_{23}$~\cite{pang2023caver} & ResNet50 & 0.927  & \textcolor[RGB]{105,189,69}{0.874} & 0.884  & \textcolor[RGB]{105,189,69}{0.043} & \textcolor[RGB]{105,189,69}{0.955} & \textcolor[RGB]{237,31,36}{0.920} & 0.919  & \textcolor[RGB]{105,189,69}{0.029} & 0.915  & 0.826  & 0.828  & \textcolor[RGB]{105,189,69}{0.044} & \textcolor[RGB]{237,31,36}{0.931} & \textcolor[RGB]{237,31,36}{0.887} & 0.871  & \textcolor[RGB]{237,31,36}{0.034} & 0.760  & \textcolor[RGB]{105,189,69}{0.663} & \textcolor[RGB]{105,189,69}{0.724} & \textcolor[RGB]{237,31,36}{0.121} & \textcolor[RGB]{237,31,36}{0.959} & \textcolor[RGB]{105,189,69}{0.899} & 0.894  & \textcolor[RGB]{105,189,69}{0.022} & 0.922  & 0.903  & 0.874  & 0.032  & 67    & 212.83  & 218.64  \bigstrut[b]\\
			\hline
			Ours  & VGG16 & 0.912  & 0.836  & 0.863  & 0.055  & 0.939  & 0.890  & 0.906  & 0.039  & 0.901  & 0.800  & 0.828  & 0.047  & 0.921  & 0.861  & 0.876  & 0.042  & 0.748  & 0.612  & 0.688  & 0.143  & 0.954  & 0.891  & 0.896  & 0.024  & 0.918  & 0.894  & 0.904  & 0.036  & 40    & 398.05  & 552.10  \bigstrut[t]\\
			Ours  & ResNet50 & 0.919  & 0.850  & 0.878  & 0.051  & 0.951  & 0.911  & 0.923  & 0.031  & \textcolor[RGB]{105,189,69}{0.919} & \textcolor[RGB]{237,31,36}{0.842} & \textcolor[RGB]{237,31,36}{0.865} & \textcolor[RGB]{237,31,36}{0.041} & 0.927  & 0.870  & 0.886  & 0.040  & \textcolor[RGB]{105,189,69}{0.763} & 0.653  & 0.719  & 0.133  & 0.955  & 0.894  & 0.896  & 0.024  & 0.918  & 0.897  & 0.906  & 0.033  & 50    & 451.80  & 137.68  \\
			Ours  & Res2Net50 & \textcolor[RGB]{237,31,36}{0.937} & \textcolor[RGB]{237,31,36}{0.883} & \textcolor[RGB]{237,31,36}{0.902} & \textcolor[RGB]{237,31,36}{0.041} & \textcolor[RGB]{237,31,36}{0.957} & \textcolor[RGB]{105,189,69}{0.919} & \textcolor[RGB]{237,31,36}{0.930} & \textcolor[RGB]{237,31,36}{0.027} & \textcolor[RGB]{237,31,36}{0.922} & \textcolor[RGB]{105,189,69}{0.840} & 0.860  & \textcolor[RGB]{237,31,36}{0.041} & \textcolor[RGB]{105,189,69}{0.930} & \textcolor[RGB]{105,189,69}{0.882} & \textcolor[RGB]{237,31,36}{0.896} & \textcolor[RGB]{105,189,69}{0.037} & 0.757  & \textcolor[RGB]{237,31,36}{0.664} & \textcolor[RGB]{237,31,36}{0.727} & \textcolor[RGB]{105,189,69}{0.128} & \textcolor[RGB]{105,189,69}{0.958} & \textcolor[RGB]{237,31,36}{0.902} & \textcolor[RGB]{237,31,36}{0.905} & \textcolor[RGB]{237,31,36}{0.021} & 0.924  & \textcolor[RGB]{237,31,36}{0.910} & \textcolor[RGB]{237,31,36}{0.919} & \textcolor[RGB]{237,31,36}{0.028} & 45    & 453.03  & 139.73  \bigstrut[b]\\
			\hline
		\end{tabular}%
	}
	\label{tab:compare2}%
\end{table*}%

\section{Experiment}
\label{sec:Experiments}

\begin{table*}[t]
	\centering
	\caption{Quantitative Comparison of E-measure (${E_\xi}$), Weighed 
		F-measure 
		(${F_\beta ^\omega}$), F-measure 
		(${F_\beta}$), Mean Absolute Error 
		($MAE$) and Frame-Per-Second (FPS) with 14 
		Different RGBT Methods on Three 
		Testing Datasets, Including VT5000~\cite{tu2020rgbt}, 
		VT1000~\cite{tu2019rgb} and VT821~\cite{tang2019rgbt}. The Best, Second 
		Best and Third Best 
		Results Are 
		Marked with \textcolor[RGB]{237,31,36}{Red} and \textcolor[RGB]{105,189,69}{Green}, Respectively}
	\setlength{\tabcolsep}{8pt}
	\renewcommand\arraystretch{1.25}
	\resizebox{1\textwidth}{!}{\small
		\begin{tabular}{c|c|cccc|cccc|cccc|c|c|c}
			\hline
			\multirow{2}[4]{*}{Methods} & \multirow{2}[4]{*}{Backbone} & \multicolumn{4}{c|}{VT5000~\cite{tu2020rgbt}}   & \multicolumn{4}{c|}{VT1000~\cite{tu2019rgb}}   & \multicolumn{4}{c|}{VT821~\cite{tang2019rgbt}} & \multirow{2}[4]{*}{FPS $\uparrow$} & \multirow{2}[4]{*}{Param $\downarrow$} & \multirow{2}[4]{*}{GFLOPs $\downarrow$} \bigstrut\\
			\cline{3-14}          &       & ${E_\xi}$ \hfill$\uparrow$     & ${F_\beta ^\omega}$ \hfill$\uparrow$    & ${F_\beta}$ \hfill$\uparrow$     & $MAE$ \hfill$\downarrow$   & ${E_\xi}$ \hfill$\uparrow$     & ${F_\beta ^\omega}$ \hfill$\uparrow$    & ${F_\beta}$ \hfill$\uparrow$     & $MAE$ \hfill$\downarrow$   & ${E_\xi}$ \hfill$\uparrow$     & ${F_\beta ^\omega}$ \hfill$\uparrow$    & ${F_\beta}$ \hfill$\uparrow$     & $MAE$ \hfill$\downarrow$ &       &       &       \bigstrut\\
			\hline
			MTMR$_{18}$~\cite{wang2018rgb}  & -     & 0.795 & 0.397 & 0.595 & 0.114 & 0.836 & 0.485 & 0.715 & 0.119 & 0.815 & 0.462 & 0.662 & 0.108 & -     & -     & -     \bigstrut[t]\\
			M3S-NIR$_{19}$~\cite{tu2019m3s} & -     & 0.780  & 0.327 & 0.575 & 0.168 & 0.827 & 0.463 & 0.717 & 0.145 & 0.859 & 0.407 & 0.734 & 0.407 & -     & -     & -     \\
			SGDL$_{19}$~\cite{tu2019rgb} & -     & 0.824 & 0.559 & 0.672 & 0.089 & 0.856 & 0.652 & 0.764 & 0.090  & 0.847 & 0.583 & 0.730  & 0.085 & -     & -     & -     \\
			ADF$_{20}$~\cite{tu2020rgbt} & VGG16 & 0.891 & 0.722 & 0.778 & 0.048 & 0.921 & 0.804 & 0.847 & 0.034 & 0.842 & 0.627 & 0.077 & 0.716 & 27    & 254.67 & 191.45 \\
			MIDD$_{21}$~\cite{tu2021multi} & VGG16 & 0.897 & 0.763 & 0.801 & 0.043 & 0.933 & 0.856 & 0.882 & 0.027 & 0.895 & 0.760  & 0.804 & 0.045 & 33    & 200   & 216.72 \\
			CSRNet$_{21}$~\cite{huo2021efficient} & ESPNetV2 & 0.905 & 0.796 & 0.811 & 0.042 & 0.925 & 0.878 & 0.877 & 0.024 & 0.909 & 0.821 & 0.831 & 0.038 & -     & -     & -     \\
			CGFNet$_{21}$~\cite{wang2021cgfnet} & VGG16 & 0.922 & 0.831 & \textcolor[RGB]{105,189,69}{0.851} & 0.035 & 0.944 & 0.900   & \textcolor[RGB]{105,189,69}{0.906} & 0.023 & 0.912 & \textcolor[RGB]{105,189,69}{0.829} & \textcolor[RGB]{237,31,36}{0.845} & 0.038 & 18    & 266.73 & 347.78 \\
			MMNet$_{21}$~\cite{gao2021unified} & Res2Net50 & 0.887 & 0.770  & 0.780  & 0.043 & 0.923 & 0.863 & 0.861 & 0.027 & 0.892 & 0.783 & 0.794 & 0.040 & -     & -     & -     \\
			ECFFNet$_{21}$~\cite{zhou2021ecffnet} & ResNet34 & 0.906 & 0.802 & 0.807 & 0.038 & 0.93  & 0.885 & 0.876 & 0.021 & 0.902 & 0.801 & 0.810  & \textcolor[RGB]{105,189,69}{0.034} & -     & -     & -     \\
			MIA-DPD$_{22}$~\cite{liang2022multi} & ResNet50 & 0.893 & 0.780  & 0.793 & 0.040  & 0.926 & 0.864 & 0.868 & 0.025 & 0.850  & 0.720  & 0.741 & 0.070 & -     & -     & -     \\
			OSRNet$_{22}$~\cite{huo2022real} & ResNet50 & 0.908 & 0.807 & 0.823 & 0.040  & 0.935 & 0.891 & 0.892 & 0.022 & 0.896 & 0.801 & 0.814 & 0.043 & \textcolor[RGB]{105,189,69}{142} & \textcolor[RGB]{105,189,69}{59.67} & \textcolor[RGB]{105,189,69}{51.27} \\
			DCNet$_{22}$~\cite{tu2022weakly} & VGG16 & 0.920  & 0.819 & 0.847 & 0.035 & \textcolor[RGB]{237,31,36}{0.948} & 0.902 & \textcolor[RGB]{237,31,36}{0.911} & 0.021 & 0.912 & 0.823 & 0.841 & \textcolor[RGB]{237,31,36}{0.033} & 43    & 91.57 & 207.21 \\
			LSNet$_{23}$~\cite{zhou2023lsnet} & MobileNetV2 & 0.915 & 0.806 & 0.825 & 0.037 & 0.935 & 0.887 & 0.885 & 0.023 & 0.911 & 0.809 & 0.825 & \textcolor[RGB]{237,31,36}{0.033} & \textcolor[RGB]{237,31,36}{314} & \textcolor[RGB]{237,31,36}{17.41} & \textcolor[RGB]{237,31,36}{3.04} \\
			CAVER$_{23}$~\cite{pang2023caver} & ResNet50 & 0.924 & \textcolor[RGB]{105,189,69}{0.835} & 0.841 & \textcolor[RGB]{105,189,69}{0.032} & \textcolor[RGB]{105,189,69}{0.945} & \textcolor[RGB]{237,31,36}{0.909} & 0.903 & \textcolor[RGB]{237,31,36}{0.017} & \textcolor[RGB]{237,31,36}{0.919} & \textcolor[RGB]{237,31,36}{0.835} & 0.839 & \textcolor[RGB]{237,31,36}{0.033} & 67    & 451.8 & 137.68 \bigstrut[b]\\
			\hline
			Ours & VGG16 & 0.917 & 0.825 & 0.838 & 0.035 & 0.939 & 0.899 & 0.897 & 0.021 & 0.904 & 0.810  & 0.824 & 0.038 & 37    & 398.05 & 552.1 \bigstrut[t]\\
			Ours & ResNet50 & \textcolor[RGB]{105,189,69}{0.925} & 0.829 & 0.847 & \textcolor[RGB]{105,189,69}{0.032} & 0.944 & 0.900   & 0.902 & 0.019 & 0.910  & 0.802 & 0.831 & \textcolor[RGB]{105,189,69}{0.034} & 49    & 451.8 & 137.68 \\
			Ours & Res2Net50 & \textcolor[RGB]{237,31,36}{0.931} & \textcolor[RGB]{237,31,36}{0.841} & \textcolor[RGB]{237,31,36}{0.857} & \textcolor[RGB]{237,31,36}{0.030} & \textcolor[RGB]{105,189,69}{0.945} & \textcolor[RGB]{105,189,69}{0.905} & 0.905 & \textcolor[RGB]{105,189,69}{0.018} & \textcolor[RGB]{105,189,69}{0.915} & 0.817 & \textcolor[RGB]{105,189,69}{0.843} & \textcolor[RGB]{105,189,69}{0.034} & 43    & 453.03 & 139.73 \bigstrut[b]\\
			\hline
		\end{tabular}%
	}
	\label{tab:compare1}%
\end{table*}%

\begin{figure*}[t]
	\centering
	\includegraphics[width=\linewidth]{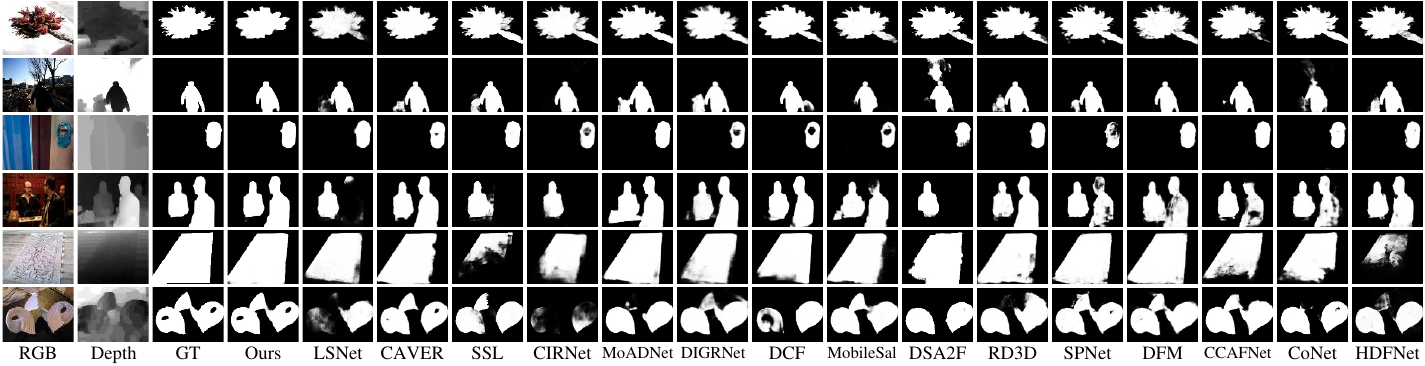}
	\caption{Qualitative visual comparisons of the proposed 
		method with other RGBD methods in diverse challenges. From the 
		${1^{st}}$ 
		row to the ${5^{th}}$ row, they are mainly challenged by depth 
		ambiguity, 
		low illumination, center bias, number variation and image clutter 
		respectively. The ${6^{th}}$ shows the the case of hollow object.}
	\label{fig::visualCompare_RGBD}
\end{figure*}

\begin{figure}[htbp]
	\centering
	\includegraphics[width=\columnwidth]{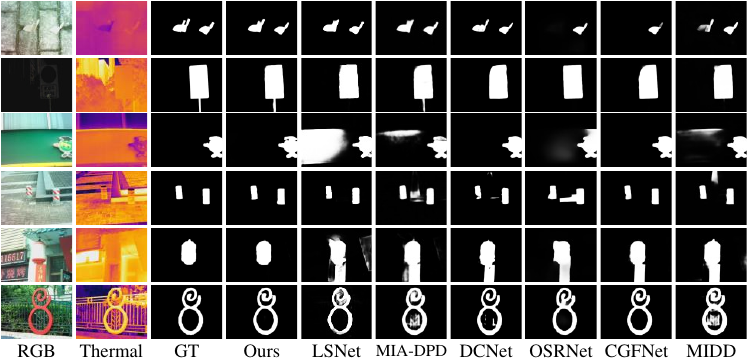}
	\caption{Qualitative visual comparisons of the proposed 
		method with other RGBT methods in diverse challenges. From the 
		${1^{st}}$ 
		row to the ${5^{th}}$ row, they are mainly challenged by thermal 
		crossover, low illumination, center bias, number variation and 
		image 
		clutter 
		respectively. The ${6^{th}}$ shows the case of hollow object.}
	\label{fig::visualCompare_RGBT}
\end{figure}

\subsection{Experimental Setup}
\textit{1) Datasets:}
We conduct RGBD experiments on seven challenging RGBD 
datasets, 
including DUT-RGBD~\cite{piao2019depth} (i.e., DUT), 
NJUD~\cite{ju2014depth}, 
NLPR~\cite{peng2014rgbd}, SIP~\cite{fan2020rethinking}, 
SSD~\cite{zhu2017three}, STERE~\cite{niu2012leveraging}, and ReDWeb-S~\cite{liu2021learning}. 
Following~\cite{sun2021deep}, we take a collection 
of 700 samples 
from NLPR, 1485 
samples from NJUD and 800 samples from DUT-RGBD as our RGBD 
training set 
and all the remaining image pairs as testing set.

The following RGBT experiments are conducted on three existing RGBT 
datasets, including VT821~\cite{tang2019rgbt}, 
VT1000~\cite{tu2019rgb} and VT5000~\cite{tu2020rgbt}. 
For a fair comparison, we train our network 
with the training set in VT5000, and take all image pairs in VT821, VT1000 
and 
the rest image pairs in VT5000 as our testing set.

\textit{2) Evaluation Metrics:}
To assess the performance of different approaches, we employ four 
widely 
used evaluation metrics, including 
E-measure (${E_\xi}$)~\cite{fan2018enhanced}, weighed F-measure 
(${F_\beta ^\omega}$)~\cite{margolin2014evaluate}, F-measure 
(${F_\beta}$)~\cite{achanta2009frequency} and mean absolute error 
($MAE$)~\cite{perazzi2012saliency}, and three efficiency evaluation metric, including FPS (frame-per-second), parameters(M) and GFLOPs. 
Specifically, E-measure is the enhanced 
alignment measure, 
which takes both of image-level mean values and pixel-level 
values into consideration. 
F-measure is a weighted combination of precision and recall. 
Mean absolute 
error accesses the difference between the prediction and its	
ground truth. 
FPS is the inference speed 
calculated in test phase with a single GPU. 
The parameters(M) of the model reflect the size of the memory occupied by the method. GFLOPs represents the amount of computation.

\textit{3) Training Details:}
Our network is trained on Pytorch in a 
workspace with two RTX 3090 GPUs. The encoder is composed of 
two identical 
Res2Net-50 networks, which are pre-trained 
on 	
ImageNet. We 
resize each of RGB image and 
thermal image or depth image into 352 × 352 for later training and 
testing. In the 
optimization stage, we 
apply the Adam algorithm with 5e-5 learning rate to train the RGBD and 
RGBT SOD models. We set 
the batch size as 16, epoch as 80. The proposed network works well in an 
end-to-end manner during 
the training and the testing stages. 
\subsection{Comparison with the State-of-the-arts}

\textit{1) Compared Methods:}
We compare our RGBD model with 17 deep learning-based state-of-the-art RGBD 
SOD 
methods, including HDFNet~\cite{HDFNet-ECCV2020}, 
CoNet~\cite{ji2020accurate}, 
CCAFNet~\cite{zhou2021ccafnet}, 
CDNet~\cite{jin2021cdnet}, 
HAINet~\cite{li2021hierarchical}, 
DFM~\cite{zhang2021depth}, 
SPNet~\cite{zhou2021specificity}, RD3D~\cite{chen2021rgb}, 
DSA2F~\cite{Sun2021DeepRS}, DCF~\cite{Ji_2021_DCF}, 
MobileSal~\cite{wu2021mobilesal}, SSL~\cite{zhao2022self}, 
DIGRNet~\cite{cheng2022depth}, CIRNet~\cite{cong2022cir}, MoADNet~\cite{jin2022moadnet}, LSNet~\cite{zhou2023lsnet}, and CAVER~\cite{pang2023caver}.
For our RGBT model, we make a comparison with 14 RGBT SOD 
methods, including MTMR~\cite{wang2018rgb}, M3S-NIR~\cite{tu2019m3s}, 
SGDL~\cite{tu2019rgb}, ADF~\cite{tu2020rgbt}, MIDD~\cite{tu2021multi}, 
CSRNet~\cite{huo2021efficient}, CGFNet~\cite{wang2021cgfnet}, 
MMNet~\cite{gao2021unified}, ECFFNet~\cite{zhou2021ecffnet}, 
MIA-DPD~\cite{liang2022multi}, OSRNet~\cite{huo2022real}, 
DCNet~\cite{tu2022weakly}, LSNet~\cite{zhou2023lsnet}, and CAVER~\cite{pang2023caver}. 
To make a fair comparison, we use the results published by the 
authors or by running their published code with default parameters.  

\begin{table}[t]
	\centering
	\caption{Performance Comparison Results of E-Measure (${E_\xi }$), F-Measure 
		(${F_\beta }$), and Mean Absolute Error ($MAE$) with 4 Transformer Backbone Based Methods on VT5000~\cite{tu2020rgbt} Test Set for RGBT Models and NJUD~\cite{ju2014depth} Test Set for RGBD Models. The Best Results Are Marked with \textbf{Bold}.}
	\renewcommand\arraystretch{1.5}
	\resizebox{1\columnwidth}{!}{\Huge
		\begin{tabular}{c|c|ccc|ccc|ccc}
			\hline
			\multirow{2}[4]{*}{Methods} & \multirow{2}[4]{*}{Backbone} & \multicolumn{3}{c|}{VT5000~\cite{tu2020rgbt}} & \multicolumn{3}{c|}{NJUD~\cite{ju2014depth}} & \multicolumn{3}{c}{STERE~\cite{niu2012leveraging}} \bigstrut\\
			\cline{3-11}          &       & ${E_\xi }$ $\uparrow$    & ${F_\beta }$ $\uparrow$     & $MAE$ $\downarrow$   & ${E_\xi }$ $\uparrow$    & ${F_\beta }$ $\uparrow$     & $MAE$ $\downarrow$   & ${E_\xi }$ $\uparrow$    & ${F_\beta }$ $\uparrow$     & $MAE$ $\downarrow$ \bigstrut\\
			\hline
			VST$_{21}$~\cite{liu2021visual} & T2T-ViT & -     & -     & -     & 0.913  & 0.856  & 0.035  & 0.916  & 0.844  & 0.038  \bigstrut[t]\\
			SwinNet$_{22}$~\cite{liu2021swinnet} & Swin Transformer & 0.942  & 0.865  & 0.026  & 0.933  & 0.921  & 0.027  & 0.929  & 0.892  & 0.033  \\
			HRTransNet$_{23}$~\cite{tang2022hrtransnet} & HRFormer & \textbf{0.945} & 0.871  & 0.025  & 0.929  & 0.919  & 0.029  & 0.929  & \textbf{0.902} & \textbf{0.030} \\
			CAVER$_{23}$~\cite{pang2023caver} & ResNet50 & 0.924  & 0.841  & 0.032  & 0.922  & 0.874  & 0.032  & 0.931  & 0.871  & 0.034  \bigstrut[b]\\
			\hline
			Ours  & Swin Transformer & 0.944  & \textbf{0.875} & \textbf{0.024} & \textbf{0.936} & \textbf{0.928} & \textbf{0.025} & \textbf{0.932} & 0.900  & \textbf{0.030} \bigstrut\\
			\hline
		\end{tabular}%
	}
	\label{tab:transformer_based}%
\end{table}%

\textit{2) Quantitative Comparison:}
Table~\ref{tab:compare2} and 
Table~\ref{tab:compare1} show quantitative comparisons with RGBD SOD methods and RGBT SOD 
methods, respectively. These methods use different backbone networks. The backbone with deeper layers (i.e., ResNet50~\cite{he2016deep}, Res2Net50~\cite{gao2019res2net}) provides networks~\cite{zhou2021specificity,cheng2022depth,cong2022cir,pang2023caver} with semantically richer features, which facilitates sufficient fusion and performance improvement but brings a larger number of parameters. In contrast, backbones with shallow layers (i.e., MobileNet~\cite{sandler2018mobilenetv2,howard2019searching}, VGG16~\cite{2014Very}) have low computational complexity and fast inference speed~\cite{zhang2021depth,zhou2023lsnet,jin2022moadnet,tu2022weakly}, but at the expense of feature representation capability, resulting in inferior performance.
Since the fusion schemes in our adaptive fusion bank (AFB) are basic and simple, designed to decouple the extracted features for different challenges, we use Res2Net-50 to extract rich features. For the RGBD comparison, our LAFB with Res2Net-50 outperforms other methods on most metrics across seven datasets. Even on the RedWeb-S~\cite{liu2021learning} dataset with diverse scenes and object categories, LAFB still has superior performance, which demonstrates the effectiveness of our method.
Compared with the advanced method SPNet~\cite{zhou2021specificity}, which also uses the Res2Net-50 backbone, our method achieves improvements of 0.8\%, 1.1\%, 1.2\%, and 4.9\% on the four metrics (${E_\xi }$, ${F_\beta ^\omega}$, ${F_\beta }$ and $MAE$) of the SIP dataset, respectively. CAVER~\cite{pang2023caver} is a recent transformer-based method with strong global context modeling ability, and our method still has comparable performance against it.
By replacing the backbone with ResNet50 or VGG16, our 
model achieves overall performance comparable to other state-of-the-art methods, such as LSNet~\cite{zhou2023lsnet}, MoADNet~\cite{jin2022moadnet} and SSL~\cite{zhao2022self}. The results on GFLOPs show that the computational complexity of our method is acceptable. Similar to the advanced methods SPNet~\cite{zhou2021specificity} and DIGRNet~\cite{cheng2022depth}, there are many parameters in our method. This is mainly comes from the multiple fusion schemes in adaptive fusion bank, which also bring better performance. Table~\ref{tab:compare1} shows the performance comparison of our method on three RGBT datasets.
On the most challenging VT5000 dataset, our method achieves an average 
improvement of 2.5\% on the four evaluation metrics, comparing with the advanced method CAVER~\cite{pang2023caver}. On the VT821 and VT1000 datasets, our method is slightly inferior to CAVER, but still outperforms other CNN-based methods. This is mainly because on the relatively simple datasets, the global modeling ability of transformer is not disturbed by various challenges and can perform better. Furthermore, we conduct a comparative analysis between our method and four recent transformer backbone based methods~\cite{liu2021visual,liu2021swinnet,tang2022hrtransnet,pang2023caver}. For a fair comparison, we replace our CNN-based backbone with Swin Transformer~\cite{liu2021swin}, as employed in SwinNet~\cite{liu2021swinnet}. The results in Table~\ref{tab:transformer_based} show the overall superior performance of our method with the support of the transformer. Compared with the suboptimal method~\cite{tang2022hrtransnet}, our method achieves an average improvement of 6.7\% on the $MAE$ metric for the three datasets.

\begin{table}[t]
	\centering
	\caption{Performance Comparison Results of Max F-Measure on 5 Challenges. LI, TC, IC, SV and CB Represent the Corresponding Challenge-Annotated Data in the Testing Set of VT5000~\cite{tu2020rgbt} for RGBT methods and  DUT-RGBD~\cite{piao2019depth} for RGBD Methods, Respectively. The Best, Second 
		Best and Third Best 
		Results Are 
		Marked with \textcolor[RGB]{237,31,36}{Red} and \textcolor[RGB]{105,189,69}{Green}, Respectively}
	\renewcommand\arraystretch{1.25}
	\resizebox{1\columnwidth}{!}{
		\begin{tabular}{c|c|ccccc}
			\hline
			\multicolumn{2}{c|}{Methods} & LI    & TC    & IC    & SV    & CB \bigstrut\\
			\hline
			\multirow{15}[4]{*}{RGBT} & MTMR~\cite{wang2018rgb}  & 0.691  & 0.597  & 0.585  & 0.661  & 0.599  \bigstrut[t]\\
			& M3S-NIR~\cite{tu2019m3s} & 0.700  & 0.608  & 0.567  & 0.633  & 0.588  \\
			& SGDL~\cite{tu2019rgb}  & 0.835  & 0.711  & 0.674  & 0.746  & 0.716  \\
			& ADF~\cite{tu2020rgbt}   & 0.868  & 0.851  & 0.833  & 0.870  & 0.859  \\
			& MIDD~\cite{tu2021multi}  & 0.894  & 0.866  & 0.841  & 0.875  & 0.871  \\
			& CSRNet~\cite{huo2021efficient} & 0.873 & 0.864 & 0.811 & 0.867 & 0.847 \\
			& CGFNet~\cite{wang2021cgfnet} & 0.891  & \textcolor[RGB]{105,189,69}{0.896} & \textcolor[RGB]{105,189,69}{0.864} & 0.886  & 0.883  \\
			& MMNet~\cite{gao2021unified} & 0.862  & 0.856  & 0.820  & 0.864  & 0.858  \\
			& ECFFNet~\cite{zhou2021ecffnet} & 0.870  & 0.872  & 0.835  & 0.880  & 0.870  \\
			& MIA-DPD~\cite{liang2022multi} & 0.892  & 0.889  & 0.852  & 0.888  & 0.882  \\
			& OSRNet~\cite{huo2022real} & 0.873  & 0.842  & 0.833  & 0.880  & 0.866  \\
			& DCNet~\cite{tu2022weakly} & 0.880  & 0.877  & 0.838  & 0.880  & 0.869  \\
			& LSNet~\cite{zhou2023lsnet} & \textcolor[RGB]{105,189,69}{0.898} & 0.865  & 0.848  & 0.881  & 0.869  \\
			& CAVER~\cite{pang2023caver} & \textcolor[RGB]{105,189,69}{0.898} & 0.877  & 0.858  & \textcolor[RGB]{105,189,69}{0.894} & \textcolor[RGB]{105,189,69}{0.890} \bigstrut[b]\\
			\cline{2-7}          & Ours LAFB & \textcolor[RGB]{237,31,36}{0.906} & \textcolor[RGB]{237,31,36}{0.900} & \textcolor[RGB]{237,31,36}{0.870} & \textcolor[RGB]{237,31,36}{0.896} & \textcolor[RGB]{237,31,36}{0.897} \bigstrut\\
			\hline
			\multirow{12}[4]{*}{RGBD} & SPNet~\cite{zhou2021specificity} & 0.873  & 0.864  & 0.864  & 0.869  & 0.867  \bigstrut[t]\\
			& RD3D~\cite{chen2021rgb}  & 0.939  & 0.932  & 0.932  & 0.944  & 0.942  \\
			& DSA2F~\cite{Sun2021DeepRS} & 0.935  & 0.929  & \textcolor[RGB]{105,189,69}{0.946} & 0.936  & 0.937  \\
			& DCF~\cite{Ji_2021_DCF}   & 0.938  & 0.935  & 0.934  & 0.937  & 0.933  \\
			& MobileSal~\cite{wu2021mobilesal} & 0.922  & 0.916  & 0.940  & 0.927  & 0.930  \\
			& SSL~\cite{zhao2022self}   & 0.913  & 0.880  & 0.907  & 0.907  & 0.915  \\
			& DIGRNet~\cite{cheng2022depth} & 0.937  & \textcolor[RGB]{105,189,69}{0.942} & 0.937  & 0.944  & 0.943  \\
			& CIRNet~\cite{cong2022cir} & \textcolor[RGB]{237,31,36}{0.947} & 0.939  & 0.943  & 0.944  & 0.941  \\
			& MoADNet~\cite{jin2022moadnet} & \textcolor[RGB]{105,189,69}{0.944} & 0.928  & 0.930  & 0.941  & 0.944  \\
			& LSNet~\cite{zhou2023lsnet} & 0.883  & 0.870  & 0.846  & 0.866  & 0.862  \\
			& CAVER~\cite{pang2023caver} & \textcolor[RGB]{237,31,36}{0.947} & \textcolor[RGB]{105,189,69}{0.942} & 0.942  & \textcolor[RGB]{105,189,69}{0.948} & \textcolor[RGB]{105,189,69}{0.947} \bigstrut[b]\\
			\cline{2-7}          & Ours LAFB & \textcolor[RGB]{237,31,36}{0.947} & \textcolor[RGB]{237,31,36}{0.946} & \textcolor[RGB]{237,31,36}{0.956} & \textcolor[RGB]{237,31,36}{0.949} & \textcolor[RGB]{237,31,36}{0.948} \bigstrut\\
			\hline
		\end{tabular}%
	}
	\label{tab:challenge}%
\end{table}%

\begin{table*}[t]
	\centering
	\caption{Ablation Studies on RGBD Datasets, Including 
		DUT-RGBD~\cite{piao2019depth} Test Set, 
		NJUD~\cite{ju2014depth} Test Set and STERE~\cite{niu2012leveraging} 
		Dataset. \emph{w/o} Means to Discard the Corresponding Component. The Best 
		Results Are Marked with \textbf{Bold}. AFB: Adaptive Fusion Bank, 
		AEM: 
		Adaptive Ensemble Module, and IIGM: Indirect Interactive Guidance 
		Module}
	\resizebox{1\textwidth}{!}{
		\begin{tabular}{c|cccc|cccc|cccc|c|c|c}
			\hline
			\multirow{2}[4]{*}{Models} & \multicolumn{4}{c|}{DUT~\cite{piao2019depth}}      & \multicolumn{4}{c|}{NJUD~\cite{ju2014depth}}     & \multicolumn{4}{c|}{STERE~\cite{niu2012leveraging}}    & \multirow{2}[4]{*}{FPS $\uparrow$} & \multirow{2}[4]{*}{parameters(M) $\downarrow$} & \multirow{2}[4]{*}{GFLOPs $\downarrow$} \bigstrut\\
			\cline{2-13}          & ${E_\xi }$ $\uparrow$    & 
			${F_\beta 
				^\omega}$ $\uparrow$    & ${F_\beta }$ $\uparrow$     & 
			$MAE$ $\downarrow$   & 
			${E_\xi }$ $\uparrow$    & 
			${F_\beta ^\omega}$ $\uparrow$    & ${F_\beta }$ $\uparrow$     
			& 
			$MAE$ $\downarrow$   & 
			${E_\xi }$ $\uparrow$    & 
			${F_\beta ^\omega}$ $\uparrow$    & ${F_\beta }$ $\uparrow$     
			& 
			$MAE$ $\downarrow$   &       &       &  \bigstrut\\
			\hline
			LAFB  & \textbf{0.957} & \textbf{0.919} & \textbf{0.930} & 
			\textbf{0.027} & \textbf{0.924} & \textbf{0.910} & \textbf{0.919} & 
			\textbf{0.028} & \textbf{0.930} & \textbf{0.882} & 
			\textbf{0.896} & \textbf{0.037} & 10    & 453.03  & 139.73  \bigstrut[t]\\
			w/o AFB & 0.940  & 0.885  & 0.906  & 0.041  & 0.911  & 0.870  & 0.892  & 0.043  & 0.922  & 0.853  & 0.884  & 0.045  & \textbf{12} & \textbf{190.93}  & \textbf{27.88}  \\
			w/o AEM & 0.946  & 0.904  & 0.920  & 0.033  & 0.912  & 0.885  & 0.902  & 0.038  & 0.926  & 0.868  & 0.885  & 0.041  & 11    & 420.78  & 118.15  \\
			w/o IIGM & 0.946  & 0.886  & 0.915  & 0.039  & 0.902  & 0.868  & 0.895  & 0.042  & 0.927  & 0.863  & 0.885  & 0.040  & 11    & 448.28  & 136.53  \bigstrut[b]\\
			\hline
		\end{tabular}%
	}
	\label{tab:ablation_rgbd}%
\end{table*}%
\begin{table*}[t]
	\centering
	\caption{Ablation Studies on RGBT Datasets, Including 
		VT5000~\cite{tu2020rgbt} Test Set, 
		VT1000~\cite{tu2019rgb} Dataset and VT821~\cite{tang2019rgbt}
		Dataset. \emph{w/o} Means to Discard the Corresponding Component. The 
		Best 
		Results Are Marked with \textbf{Bold}. AFB: Adaptive Fusion Bank, 
		AEM: 
		Adaptive Ensemble Module, and IIGM: Indirect Interactive Guidance 
		Module}
	\resizebox{1\textwidth}{!}{
		\begin{tabular}{c|cccc|cccc|cccc|c|c|c}
			\hline
			\multirow{2}[4]{*}{Models} & \multicolumn{4}{c|}{VT5000~\cite{tu2020rgbt}}   & \multicolumn{4}{c|}{VT1000~\cite{tu2019rgb}}   & \multicolumn{4}{c|}{VT821~\cite{tang2019rgbt}}    & \multirow{2}[4]{*}{FPS $\uparrow$} & \multirow{2}[4]{*}{parameters(M) $\downarrow$} & \multirow{2}[4]{*}{GFLOPs $\downarrow$} \bigstrut\\
			\cline{2-13}          & ${E_\xi }$ $\uparrow$    & 
			${F_\beta 
				^\omega}$ $\uparrow$    & ${F_\beta }$ $\uparrow$     & 
			$MAE$ $\downarrow$   & 
			${E_\xi }$ $\uparrow$    & 
			${F_\beta ^\omega}$ $\uparrow$    & ${F_\beta }$ $\uparrow$     
			& 
			$MAE$ $\downarrow$   & 
			${E_\xi }$ $\uparrow$    & 
			${F_\beta ^\omega}$ $\uparrow$    & ${F_\beta }$ $\uparrow$     
			& 
			$MAE$ $\downarrow$   &       &       &  \bigstrut\\
			\hline
			LAFB  & \textbf{0.931} & \textbf{0.841} & \textbf{0.857} & 
			\textbf{0.030} & \textbf{0.945} & \textbf{0.905} & \textbf{0.905} & 
			\textbf{0.018} & \textbf{0.915} & \textbf{0.817} & 
			\textbf{0.843} & \textbf{0.034} & 7    & 453.03  & 139.73  \bigstrut[t]\\
			w/o AFB & 0.904  & 0.799  & 0.815  & 0.041  & 0.931  & 0.880  & 0.882  & 0.026  & 0.885  & 0.778  & 0.792  & 0.046  & \textbf{9} & \textbf{190.93}  & \textbf{27.88}  \\
			w/o AEM & 0.917  & 0.819  & 0.839  & 0.036  & 0.941  & 0.899  & 0.899  & 0.018  & 0.886  & 0.783  & 0.803  & 0.057  & 8     & 420.78  & 118.15  \\
			w/o IIGM & 0.912  & 0.815  & 0.830  & 0.037  & 0.937  & 0.893  & 0.893  & 0.021  & 0.897  & 0.792  & 0.813  & 0.040  & 7     & 448.28  & 136.53  \bigstrut[b]\\
			\hline
		\end{tabular}%
	}
	\label{tab:ablation_rgbt}%
\end{table*}%
It is worth noting 
that most of these compared methods in Table~\ref{tab:compare2} and Table~\ref{tab:compare1} only perform one of the RGBD and RGBT 
tasks, but our method is applied to both tasks and achieves impressive 
performance. Our method demonstrates its strong 
generalization ability, as it is applied to both RGBD and RGBT tasks and achieves impressive performances, mainly due to the fusion schemes designed for the challenges in multi-modal images.

\textit{3) Qualitative Comparison:}
Fig.~\ref{fig::visualCompare_RGBD} and Fig.~\ref{fig::visualCompare_RGBT} illustrate the qualitative comparisons of our method against other top-ranking methods in various challenging scenarios, including depth ambiguity or thermal crossover (the ${1^{st}}$ row), low illumination (the ${2^{nd}}$ row), center bias (the ${3^{rd}}$ row), number variation (the ${4^{th}}$ row) and image clutter (the ${5^{th}}$ row). On the one hand, our method can segment salient regions more accurately in each case, whereas other methods only handle a few challenges well. For example, CIRNet~\cite{cong2022cir} can handle the low illumination challenge (i.e., the ${2^{nd}}$ row) well but fails on other challenges, and DCNet~\cite{tu2022weakly} can handle center bias and thermal crossover challenges (i.e., the ${1^{st}}$ row and the ${3^{rd}}$ row) well but also fails on other challenges. On the other hand, though some images contain multiple challenges, our method is able to exploit the complementarity of multiple fusion schemes to tackle them well, such as the ${5^{th}}$ row in the RGBD comparison with the challenges of both image clutter and scale variation. The last row of both sets of comparisons is about the hollow object. The comparison results indicate that our method segment the foreground from background more precisely, which relies heavily on IIGM to balance semantic and detailed information.

\textit{4) Challenge-based Comparison:}
We further make a comparison with other methods on the five challenges labeled by VT5000~\cite{tu2020rgbt}. Following~\cite{tu2021multi,liao2022cross}, we compute the max F-measure score of each method on each challenge, and the results are presented in the upper part of Table~\ref{tab:challenge}. LI, TC, IC, SV and CB represent low illumination, thermal crossover, image clutter, scale variation and center bias challenges, respectively. On all challenges, our method outperforms the compared methods, which proves 
that our method can address these challenges simultaneously. The main reason is that our adaptive fusion bank can select corresponding fusion schemes to deal with challenges in different input data.
Although CAVER~\cite{pang2023caver}, LSNet~\cite{zhou2023lsnet} and CGFNet~\cite{wang2021cgfnet} achieve sub-optimal 
performance on several challenges, they fail to perform well on all challenges 
at the same time. This demonstrates that the fusion strategies they designed focus on addressing specific challenges and have difficulty generating features for addressing some other important challenges.
Note that the sub-optimal method CAVER~\cite{pang2023caver} is a recent transformer-based method with a strong ability to model global contextual information, but it still performs poorly on some challenges that require both local and global information, such as the performance on image clutter (IC) challenge. Specifically, compared to it, our method achieves performance improvements of 0.9\%, 2.6\%, 1.4\%, 0.2\% and 0.8\% on the five challenge (i.e., LI, TC, IC, SV, CB) data, respectively.
Furthermore, on the center bias (CB) challenge, our method outperforms most methods (e.g., LSNet, DCNet, OSRNet) by a large margin, even though the fusion scheme used to solve the CB challenge only consists of one convolutional block. This validates that the fusion scheme we designed is simple but effective. 
In addition, following the rules in VT5000~\cite{tu2020rgbt}, we annotate the challenge attributes of each image pair in the DUT-RGBD~\cite{piao2019depth} dataset. Based on this, we also make a comparison with recent advanced RGBD SOD methods on the five challenges, and the results are shown in the lower part of Table~\ref{tab:challenge}. It can be found that our method also achieves optimality on all challenges, which demonstrates the effectiveness of our method to deal with the challenging scenes in the RGBD SOD dataset.

\begin{table}[t]
	\centering
	\caption{Performance Comparison Results of Max F-Measure on 5 Challenges. 
		LI, TC, IC, SV and CB Represent the Corresponding Challenge-annotated 
		Data 
		in the Testing Set of VT5000~\cite{tu2020rgbt}, 
		Respectively. \emph{only} Means to Cover the Fusion Bank with Only 
		One Fusion Scheme. The Best Results Are Marked in \textbf{Bold} and 
		the Subo-ptimal Results Are \underline{Underlined}}.
	\resizebox{1\width}{!}{\scriptsize
		\begin{tabular}{c|ccccc}
			\hline
			Models & LI    & TC    & IC    & SV    & CB \bigstrut\\
			\hline
			LAFB  & \textbf{0.906} & \textbf{0.900} & \textbf{0.870} & 
			\textbf{0.896} & \textbf{0.897} \bigstrut\\
			\hline
			\emph{only} $Fu{s^{li}}$ & \underline{0.900}  & 0.866  & 0.835  & 
			0.871  & 
			0.864  
			\bigstrut[t]\\
			\emph{only} $Fu{s^{td}}$ & 0.889  & \underline{0.897}  & 0.844  & 
			0.869  & 
			0.863  \\
			\emph{only} $Fu{s^{ic}}$ & 0.869  & 0.885  & \underline{0.852}  & 
			0.878  & 
			0.874  \\
			\emph{only} $Fu{s^{sv}}$ & 0.892  & 0.867  & 0.844  & 
			\underline{0.886}  & 
			0.878  \\
			\emph{only} $Fu{s^{cb}}$ & 0.886  & 0.865  & 0.846  & 0.882  & 
			\underline{0.884}  \\
			LAFB (\emph{w/o} AFB) & 0.875  & 0.863  & 0.819  & 0.857  & 0.850  
			\bigstrut[b]\\
			\hline
		\end{tabular}%
	}
	\label{tab:ablation_branch}%
\end{table}%
\begin{figure}[t]
	\centering
	\includegraphics[width=\linewidth]{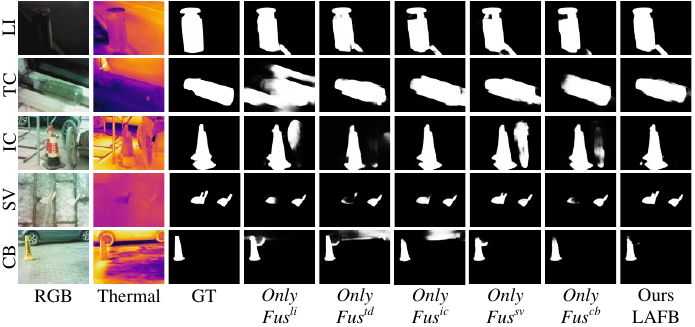}
	\caption{Visual ablation results for each single fusion scheme on the five challenges. \emph{only}: cover all fusion schemes with the corresponding only one fusion scheme. LI, TC, IC, SV and CB represent the challenges.}
	\label{fig::visual_Table6}
\end{figure}

\subsection{Ablation Study}
\label{ablation}
In order to verify the effectiveness of the main 
components in our 
framework, we carry out ablation studies to investigate their 
importance. The results on RGBD and RGBT datasets 
are shown in Table~\ref{tab:ablation_rgbd} and 
Table~\ref{tab:ablation_rgbt}, 
respectively. In these tables, \emph{w/o} means disabling the corresponding 
components and 
LAFB means our proposed complete model. We also explore the 
effectiveness of each fusion scheme in the adaptive fusion bank, and the 
results are shown in Table~\ref{tab:ablation_branch}. Table~\ref{tab:iigm} and 
Table~\ref{tab:loss} present the performance comparison results of different 
integration strategies for multi-level features and different loss settings, 
respectively.

\textit{1) Effect of adaptive fusion bank:}
To evaluate the 
effectiveness of our adaptive fusion bank (AFB), we compare the 
performance 
of our network with that of removing the AFB, denoted as 
'\emph{w/o} AFB' in the first row of Table~\ref{tab:ablation_rgbd} and 
Table~\ref{tab:ablation_rgbt}. This means that the features of the two 
modalities have no interaction for challenges. Specifically, we 
directly concatenate the 
multi-modal 
features in a channel-wise manner and feed them into the following module. 
Without the AFB, the average values on the 
four evaluation metrics (${E_\xi }$, ${F_\beta ^\omega}$, ${F_\beta }$, 
$MAE$)  
drop by 1.3\%, 3.6\%, 2.3\%, 42.4\% on the three RGBD test datasets, and 
2.6\%, 4.2\%, 4.4\%, 38.8\% on the three RGBT test datasets, 
respectively. Moreover, the results on FPS, parameters and GFLOPs show that the computational complexity of the model without AFB are significantly reduced, because AFB integrates multiple fusion branches, and is embedded into multiple layers.

To verify that each fusion scheme specializes in its corresponding challenges, we evaluate the performance of different fusion schemes on the five challenges in Table~\ref{tab:ablation_branch}. Specifically, we cover the entire fusion bank with five fusion schemes sequentially to obtain different models, and then train them successively with the entire training set of VT5000~\cite{tu2020rgbt}. Next, we compute the max F-measure score for each model on the five challenges annotated in the testing set of VT5000. The results indicate that each fusion scheme achieves better performance on the challenge that they should address, despite its simple structure. The comparison between these models and our LAFB also demonstrates that our method fully leverage the complementary benefits of these basic fusion schemes to handle different challenges effectively. In the SV column of Table~\ref{tab:ablation_branch}, $Fu{s^{cb}}$ performs better than $Fu{s^{ic}}$. The reasons for this are as follows: First, in scale variation challenge, multiple objects and large objects can easily cause objects to be off-center, which falls under the center bias (CB) challenge that is handled by $Fu{s^{cb}}$. Second, while the two convolution layers in $Fu{s^{ic}}$ bring a higher receptive field, they also lose much local and detailed information, which is preserved by $Fu{s^{cb}}$ and is important for accurately recognizing small objects under the scale variation challenge. Fig.~\ref{fig::visual_Table6} vividly illustrates the visual results of the corresponding models in Table~\ref{tab:ablation_branch} across different challenge scenes. It can be seen that models with only a single fusion scheme focus on addressing their own challenges, while failing to address all the challenges well at the same time. In contrast, our whole model is able to adaptively exploit the advantages of the corresponding fusion schemes based on different inputs to solve multiple challenges simultaneously.

Furthermore, we perform the experiments with a combination of different fusion schemes, and the results are shown in Table~\ref{tab:fusion_combination}. It can be seen that with more fusion schemes, the model has higher performance, and the whole model (i.e., LAFB) with all five fusion schemes achieves the optimal performance. In addition, compared with the removal of all fusion schemes (i.e., w/o AFB) in Table~\ref{tab:ablation_rgbd} and~\ref{tab:ablation_rgbt}, the models with fusion schemes have superior performance on all three metrics (i.e., ${E_\xi }$, ${F_\beta }$, and $MAE$). This is mainly because models with more fusion schemes can comprehensively deal with different challenges.

\begin{table}[t]
	\centering
	\caption{Performance Comparison Results of Different Combinations of Fusion Schemes on VT5000~\cite{tu2020rgbt} Test Set for the RGBT Model and NJUD~\cite{ju2014depth} Test Set for the RGBD Model. The Best Results Are Marked with \textbf{Bold}.}
	\renewcommand\arraystretch{1.3}
	\resizebox{1\columnwidth}{!}{\LARGE
		\begin{tabular}{c|ccccc|ccc|ccc}
			\hline
			\multirow{2}[4]{*}{Models} & \multicolumn{5}{c|}{Fusion schemes}   & \multicolumn{3}{c|}{VT5000~\cite{tu2020rgbt}} & \multicolumn{3}{c}{NJUD~\cite{ju2014depth}} \bigstrut\\
			\cline{2-12}          & $Fu{s^{li}}$ & $Fu{s^{td}}$ & $Fu{s^{ic}}$ & $Fu{s^{sv}}$ & $Fu{s^{cb}}$ & ${E_\xi }$ $\uparrow$    & ${F_\beta }$ $\uparrow$     & $MAE$ $\downarrow$   & ${E_\xi }$ $\uparrow$    & ${F_\beta }$ $\uparrow$     & $MAE$ $\downarrow$ \bigstrut\\
			\hline
			(a)   & \checkmark     &       & \checkmark     &       &       & 0.917  & 0.838  & 0.035  & 0.917  & 0.902  & 0.038  \bigstrut[t]\\
			(b)   &       & \checkmark     & \checkmark     &       &       & 0.917  & 0.839  & 0.036  & 0.913  & 0.898  & 0.038  \\
			(c)   & \checkmark     & \checkmark     & \checkmark     &       &       & 0.920  & 0.842  & 0.033  & 0.915  & 0.905  & 0.034  \\
			(d)   & \checkmark     & \checkmark     & \checkmark     & \checkmark     &       & 0.928  & 0.852  & 0.031  & 0.919  & 0.912  & 0.031  \\
			LAFB  & \checkmark     & \checkmark     & \checkmark     & \checkmark     & \checkmark     & \textbf{0.931} & \textbf{0.857} & \textbf{0.030} & \textbf{0.924} & \textbf{0.919} & \textbf{0.028} \bigstrut[b]\\
			\hline
		\end{tabular}%
	}
	\label{tab:fusion_combination}%
\end{table}%
\textit{2) Effect of adaptive Ensemble Module:}
To verify the effectiveness of AEM, we replace it with channel-wise 
concatenation, which means that the fusion bank cannot select fusion schemes 
adaptively. The results are shown 
in the third row of both Table~\ref{tab:ablation_rgbd} and 
Table~\ref{tab:ablation_rgbt}. Without AEM, the average value of the $MAE$ metric drops by an average of 22.9\% on the three RGBD datasets. Comparison with the results of our LAFB 
illustrates that AEM further improves the performance based on different 
fusion schemes. This is mainly because AEM enhances the output features of corresponding
fusion schemes that can deal with the input challenges. 

\begin{table}[t]
	\centering
	\caption{Performance Comparison Results of Different Integration Strategies 
		of the Indirect Interactive Guidance Module on VT5000~\cite{tu2020rgbt} 
		Test 
		Set for the RGBT Model and 
		NJUD~\cite{ju2014depth} Test Set for the RGBD Model. The Best Results Are Marked with \textbf{Bold}. Top-down: Integrating Features Layer by Layer, and Direct: 
		Integrating All Four-Layer Features}
	\renewcommand\arraystretch{1.1}
	\resizebox{1\width}{!}{\scriptsize
		\begin{tabular}{c|l|ccc|ccc}
			\hline
			\multicolumn{2}{c|}{\multirow{2}[4]{*}{Models}} & \multicolumn{3}{c|}{VT5000~\cite{tu2020rgbt}} & \multicolumn{3}{c}{NJUD~\cite{ju2014depth}} \bigstrut\\
			\cline{3-8}    \multicolumn{2}{c|}{} & \multicolumn{1}{c}{${E_\xi }$ $\uparrow$} & \multicolumn{1}{c}{${F_\beta }$ 
				$\uparrow$} & \multicolumn{1}{c|}{$MAE$ $\downarrow$} & ${E_\xi }$ $\uparrow$    & ${F_\beta }$ 
			$\uparrow$     & $MAE$ $\downarrow$ \bigstrut\\
			\hline
			\multicolumn{2}{c|}{LAFB} & \textbf{0.931} & \textbf{0.857} & \textbf{0.030} & \textbf{0.924} & \textbf{0.919} & \textbf{0.028} \bigstrut\\
			\hline
			\multirow{3}[2]{*}{\begin{sideways}Top-down\end{sideways}} & (a)   & 0.918  & 0.835  & 0.034  & 0.910  & 0.905  & 0.037  \bigstrut[t]\\
			& (b)~\cite{zhou2021specificity} & 0.920  & 0.843  & 0.035  & 0.911  & 0.905  & 0.034  \\
			& (c)~\cite{Ji_2021_DCF} & 0.922  & 0.848  & 0.033  & 0.915  & 0.906  & 0.035  \bigstrut[b]\\
			\hline
			\multirow{3}[2]{*}{\begin{sideways}Direct\end{sideways}} & (d)   & 0.926  & 0.852  & 0.032  & 0.922  & 0.915  & 0.031  \bigstrut[t]\\
			& (e)~\cite{zhang2020feature} & 0.924  & 0.845  & 0.031  & 0.917  & 0.913  & 0.032  \\
			& (f)~\cite{Sun2021DeepRS}  & 0.923  & 0.845  & 0.032  & 0.923  & 0.910  & 0.032  \bigstrut[b]\\
			\hline
		\end{tabular}%
		
	}
	\label{tab:iigm}%
\end{table}%
\begin{figure}[t]
	\centering
	\includegraphics[width=\linewidth]{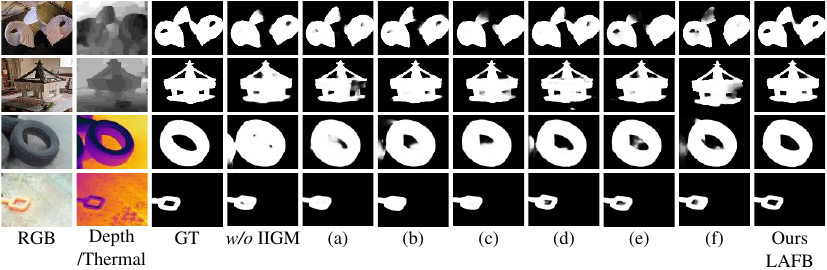}
	\caption{Saliency maps for different integration approaches of multi-level 
		features.\emph{w/o} IIGM: removing IIGM in the network, Direct 
		integration: integrating two-layer features, and Full integration: 
		integrating all four-layer features.}
	\label{fig::visual_IIGM}
\end{figure}
\textit{3) Effect of indirect interactive guidance module:}
To verify the effectiveness of the indirect interactive guidance module 
(IIGM), we conduct an ablation study by removing the IIGM 
and feeding the hierarchical outputs of AFB directly into the 
decoder to generate predictions. The model is denoted as "\emph{w/o} IIGM" in Table 
~\ref{tab:ablation_rgbd} 
and Table 
~\ref{tab:ablation_rgbt}. The results show that IIGM improves the performance of our model on all evaluation metrics 
through the indirect integration and guidance of multi-level features. Without IIGM, the average value on $MAE$ metric drops by an average of 34.2\% on the three RGBD datasets. Furthermore, the impact of IIGM on the computational complexity is negligible, as shown in the results on FPS, parameters, and GFLOPs.

Moreover, we conduct experiments to compare the IIGM with top-down and direct integration in Table~\ref{tab:iigm}. For the top-down integration, we first directly fuse features layer by layer through a convolutional block, with the results reported in Table~\ref{tab:iigm} (a). To further validate the effectiveness of the IIGM, we replace the IIGM with other top-down integration methods~\cite{zhou2021specificity,Ji_2021_DCF}, and the results are shown in Table~\ref{tab:iigm} (b) and (c). By comparing these methods with the first row (i.e., LAFB) in Table~\ref{tab:iigm}, we observe 
that the top-down integration has lower performance on all evaluation metrics. The limited number of adjacent layers restricts the diversity of features and prevents the full integration of multi-level features. For direct integration, we first fuse all four-layer features directly, with the results shown in Table~\ref{tab:iigm} (d). In addition, we also replace the IIGM with other direct integration methods~\cite{Sun2021DeepRS,zhang2020feature}, the results are denoted as (e) and (f) shown in Table~\ref{tab:iigm}. These direct integration methods neglect the differences between high-level and low-level features and mix all features without specific guidance, resulting in sub-optimal performance. Nonetheless, integrating multi-layer features in Table~\ref{tab:iigm} all yield better results than "\emph{w/o} IIGM" in Table~\ref{tab:ablation_rgbd} and Table~\ref{tab:ablation_rgbt}, proving that multi-layer fusion can aid in accurately detecting salient regions. Fig.~\ref{fig::visual_IIGM} also illustrates the saliency maps for several representative examples of hollow objects. Our method accurately locates the salient regions while suppressing the non-salient regions in the objects. For example, in the first and second rows of Fig.~\ref{fig::visual_IIGM}, although the hollowed-out background area is small, our method is still able to suppress it. This is because our IIGM can fully integrate semantic and detailed information in multi-level features.

\begin{table}[t]
	\centering
	\caption{Performance Comparison Results of Different Loss Settings on 
		VT5000~\cite{tu2020rgbt} Test 
		Set for the RGBT Model and 
		NJUD~\cite{ju2014depth} Test Set for the RGBD Model. The 
		Best 
		Results Are Marked with \textbf{Bold}. $\lambda _i$, ${\ell _s}$ and 
		${\ell _d}$ Represent Weights Parameters, Smoothness Loss and Dice 
		Loss, Respectively}
	\renewcommand\arraystretch{1.1}
	\resizebox{1\columnwidth}{!}{\Large
		\begin{tabular}{c|cccccc|ccc|ccc}
			\hline
			\multirow{2}[4]{*}{Models} & \multicolumn{6}{c|}{Loss 
				settings}            & 
			\multicolumn{3}{c|}{VT5000~\cite{tu2020rgbt}} & 
			\multicolumn{3}{c}{NJUD~\cite{ju2014depth}} \bigstrut\\
			\cline{2-13}          & $\lambda _2$    & 
			\multicolumn{1}{l}{$\lambda _3$} & 
			\multicolumn{1}{l}{$\lambda _4$} & \multicolumn{1}{l}{$\lambda _5$} 
			& 
			\multicolumn{1}{l}{${\ell _s}$} & \multicolumn{1}{l|}{${\ell _d}$} 
			& ${E_\xi }$ $\uparrow$    & ${F_\beta }$ $\uparrow$     & 
			$MAE$ $\downarrow$   & ${E_\xi }$ $\uparrow$    & ${F_\beta }$ 
			$\uparrow$     & 
			$MAE$ $\downarrow$ \bigstrut\\
			\hline
			LAFB  & 1.0   & 0.8   & 0.6   & 0.5   & \checkmark   & \checkmark   
			& \textbf{0.931} & \textbf{0.857} & \textbf{0.030} & 
			\textbf{0.924} & \textbf{0.919} & \textbf{0.028} \bigstrut[t]\\
			(a)   & 1.0   & 1.0   & 1.0   & 1.0   & \checkmark   & \checkmark   
			& 0.920  & 0.845  & 0.033  & 0.920  & 0.910  & 0.033  \\
			(b)   & 1.0   & 0.0   & 0.0   & 0.0   & \checkmark   & \checkmark   
			& 0.928  & 0.850  & 0.032  & 0.918  & 0.906  & 0.037  \\
			(c)   & 1.0   & 0.8   & 0.6   & 0.5   &    & \checkmark   & 0.925  
			& 0.842  & 0.032  & 0.920  & 0.908  & 0.033 \\
			(d)   & 1.0   & 0.8   & 0.6   & 0.5   & \checkmark   &    & 0.922  
			& 0.846  & 0.033  & 0.920  & 0.908  & 0.033  \\
			(e)   & 1.0   & 0.8   & 0.6   & 0.5   &    &    & 0.914  & 0.837  & 
			0.036  & 0.910  & 0.894  & 0.037  \bigstrut[b]\\
			\hline
		\end{tabular}%
	}
	\label{tab:loss}%
\end{table}%

\begin{figure}[t]
	\centering
	\includegraphics[width=\linewidth]{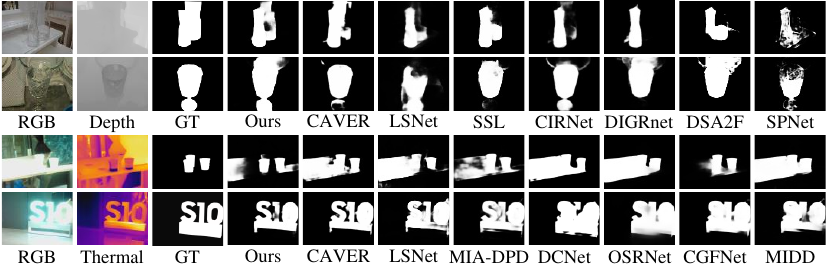}
	\caption{Visual results of some typical failure cases from RGBD and RGBT testing datasets.}
	\label{fig::failure}
\end{figure}
\textit{4) Effect of loss function:}
Our total loss (i.e., ${\ell _{total}}$) consists of the main loss 
(i.e., ${\ell _2}$, ${\ell _s}$ and ${\ell _d}$) 
and 
the auxiliary 
loss (i.e., ${\ell _3}$, ${\ell 
	_4}$ and ${\ell _5}$). The main loss 
is used to constrain 
the final 
saliency map $S_2$, and the auxiliary loss is used to 
constrain the intermediate features of the decoder to 
learn discriminative information for saliency detection. We set the weight 
parameters of the 
main loss to the default value of 1.0 to emphasize its dominance. The weight 
parameters (i.e., ${\lambda _3}$, ${\lambda _4}$ and ${\lambda _5}$) 
in the auxiliary loss are set to 0.8, 0.6 and 0.5 
in a 
decreasing manner to gradually weaken the impact of the intermediate features.

To demonstrate the effectiveness of the auxiliary loss, we first set the weight parameters (i.e., ${\lambda _3}$, ${\lambda _4}$ 
and ${\lambda _5}$) to 1.0, treating all losses 
equally. As shown in Table~\ref{tab:loss}, the comparison between this setting and our 
LAFB confirms the importance of discriminative treatment of the
auxiliary loss. Additionally, we disable the auxiliary loss by setting its weight parameters to 0. Comparing the results in Table~\ref{tab:loss} (b) with our LAFB, we observe that 
removing the 
auxiliary loss degrades the performance. Furthermore, comparing the results in (a) and (b) shows that the equal treatment of the auxiliary loss can have negative effects.

In order to verify the role of smoothness loss (i.e., ${\ell 
	_s}$) and dice loss (i.e., ${\ell _d}$) in the main loss, we remove them 
successively. The results are shown in (c) and (d) of 
Table~\ref{tab:loss}, respectively. It can be seen that the smoothness loss and 
the dice loss 
play an important role in optimizing the model. Furthermore, we remove the 
smoothness loss and the dice loss simultaneously to observe their joint 
effectiveness. As shown in (e) of 
Table~\ref{tab:loss}, the average index of three metrics (${E_\xi 
}$, ${F_\beta }$, 
$MAE$) on the two testing datasets decrease by 1.7\%, 2.6\% and 26.1\%, 
respectively. It indicates that using both ${\ell _s}$ and ${\ell _d}$ works 
better than using one of them alone.

\subsection{Failure Cases and Analyses}
\label{failure}

Although our method is able to address the main challenges in multi-modal SOD, there are still some failure cases, as shown in Fig.~\ref{fig::failure}. For the samples in the first and second rows, the salient objects are transparent, leading to a lack of visible color and texture information in the RGB modality. In this case, our model and other models struggle to accurately locate the salient region and tend to introduce noise from the background and depth map. For the samples in the third and fourth rows, overexposure causes both our model and the existing state-of-the-art ones to focus more on the exposed area, which makes it difficult to distinguish foreground and background areas. These issues remain to be resolved, and we will attempt to design a common fusion scheme to deal with the remaining challenges in future work.

\section{Conclusion}
In this paper, we propose to learn an adaptive fusion
bank (LAFB) for robust multi-modal salient object detection (MSOD). 
Considering the characteristics of different challenges, we design a 
fusion bank comprising five 
fusion schemes for the five main challenges in MSOD.
In order to further form the adaptive fusion bank (AFB), we introduce an 
adaptive 
ensemble module (AEM), which takes advantage of the complementarity of the 
fusion schemes. In addition, the proposed 
indirect interactive guidance module (IIGM) improves 
the prediction for hollow objects in MSOD by fully integrating 
high-level semantic and low-level detailed features. Experimental 
results 
on multiple multi-modal 
datasets prove the superiority of our 
LAFB and the effectiveness of the proposed modules. 

Although this paper addresses the main and important challenges in MSOD, 
there are still some difficult challenges to be addressed, such as occlusion, 
blur, 
etc. In the future, we will consider developing a common fusion scheme to deal 
with the remaining challenges and integrate it with the 
existing AFB to handle the challenges in MSOD more comprehensively.

\bibliographystyle{IEEEtran}
\bibliography{mybibfile}

\end{document}